\documentclass[sigconf]{acmart}

\usepackage{graphicx}
\usepackage{tabularx}
\usepackage[mathscr]{euscript}
\usepackage{amsmath}
\usepackage{multirow}
\usepackage{amsfonts}
\usepackage{float}
\usepackage{algorithm}
\usepackage{algpseudocode}
\newcolumntype{x}[1]{>{\centering\arraybackslash}p{#1}}
\newcolumntype{Y}{>{\centering\arraybackslash}X}
\newcommand{\minimize}{\mathop{\mathrm{minimize}}}
\usepackage{mwe}
\usepackage{subcaption}
\usepackage{balance}
\usepackage{color}
\AtBeginDocument{%
  \providecommand\BibTeX{{%
    \normalfont B\kern-0.5em{\scshape i\kern-0.25em b}\kern-0.8em\TeX}}}

\setcopyright{acmcopyright}
\copyrightyear{2020} 
\acmYear{2020} 
\setcopyright{acmcopyright}
\acmConference[KDD '20] {26th ACM SIGKDD Conference on Knowledge Discovery and Data Mining}{August 23--27, 2020}{Virtual Event, USA}
\acmBooktitle{26th ACM SIGKDD Conference on Knowledge Discovery and Data Mining (KDD '20), August 23--27, 2020, Virtual Event, USA}
\acmPrice{15.00}
\acmDOI{10.1145/XXXXXX.XXXXXX}
\acmISBN{978-1-4503-7998-4/20/08} 

\begin{document}
\fancyhead{}

%

%
\title{Connecting the Dots:  Multivariate Time Series Forecasting with Graph Neural Networks}

%

\author{Zonghan Wu}
\affiliation{%
  \institution{University of Technology Sydney}
}
\email{zonghan.wu-3@student.uts.edu.au}

\author{Shirui Pan}
\authornote{Corresponding Author.}
\affiliation{%
  \institution{Monash University}
}
\email{shirui.pan@monash.edu}

\author{Guodong Long}
\affiliation{%
  \institution{University of Technology Sydney}
}
\email{guodong.long@uts.edu.au}

\author{Jing Jiang}
\affiliation{%
  \institution{University of Technology Sydney}
}
\email{jing.jiang@uts.edu.au}

\author{Xiaojun Chang}
\affiliation{%
  \institution{Monash University}
}
\email{xiaojun.chang@monash.edu}

\author{Chengqi Zhang}
\affiliation{%
  \institution{University of Technology Sydney}
}
\email{chengqi.zhang@uts.edu.au}

\renewcommand{\shortauthors}{Wu, et al.}

\begin{abstract}
Modeling multivariate time series has long been a subject that has attracted researchers from a diverse range of fields including economics, finance, and traffic. A basic assumption behind multivariate time series forecasting is that its variables depend on one another but, upon looking closely, it's fair to say that existing methods fail to fully exploit latent spatial dependencies between pairs of variables. In recent years, meanwhile, graph neural networks (GNNs) have shown high capability in handling relational dependencies. 
GNNs require well-defined graph structures for information propagation which means they cannot be applied directly for multivariate time series where the dependencies are not known in advance. In this paper, we propose a general graph neural network framework designed specifically for multivariate time series data. Our approach automatically extracts the uni-directed relations among variables through a graph learning module, into which external knowledge like variable attributes can be easily integrated. A novel mix-hop propagation layer and a dilated inception layer are further proposed to capture the spatial and temporal dependencies within the time series. The graph learning, graph convolution, and temporal convolution modules are jointly learned in an end-to-end framework. Experimental results  
show that our proposed model outperforms the state-of-the-art baseline methods on 3 of 4 benchmark datasets and achieves on-par performance with other approaches on two traffic datasets which provide extra structural information.
\end{abstract}


%
%



\maketitle


\section{Introduction}
Modern societies have benefited from a wide range of sensors to record changes in temperature, price, traffic speed, electricity usage, and many other forms of data. Recorded time series from different sensors can form multivariate time series data and can be interlinked. For example, the rise in daily temperature may cause an increase in electricity usage. To capture systematic trends over a group of dynamically changing variables, the problem of multivariate time series forecasting has been studied for at least sixty years. It has seen tremendous applications in the domains of economics, finance, bioinformatics, and traffic.

Multivariate time series forecasting methods inherently assume interdependencies among variables. In other words, each variable depends not only on its historical values but also on other variables. However, existing methods do not exploit latent interdependencies among variables efficiently and effectively. Statistical methods, such as vector auto-regressive model (VAR) and Gaussian process model (GP), assume a linear dependency among variables. The model complexity of statistical methods grows quadratically with the number of variables. They face the problem of overfitting with a large number of variables. Recently developed deep-learning-based methods, including LSTNet \cite{lai2018modeling} and TPA-LSTM \cite{shih2019temporal}, are powerful to capture non-linear patterns. LSTNet encodes short-term local information into low dimensional vectors using 1D convolutional neural networks and decodes the vectors through a recurrent neural network. TPA-LSTM processes the inputs by a recurrent neural network and employs a convolutional neural network to calculate the attention score across multiple steps. LSTNet and TPA-LSTM do not model the pair-wise dependencies among variables explicitly, which weakens model interpretability. 

Graphs are a special form of data which describes the relationships between different entities. Recently, graph neural networks have achieved great success in handling graph data due to their permutation-invariance, local connectivity, and compositionality. By propagating information through structures, graph neural networks allow each node in a graph to be aware of its neighborhood context. Multivariate time series forecasting can be viewed naturally from a graph perspective. Variables from multivariate time series can be considered as nodes in a graph, and they are interlinked through their hidden dependency relationships. It follows that modeling multivariate time series data using graph neural networks can be a promising way to preserve their temporal trajectory while exploiting the interdependency among time series.

The most suitable type of graph neural networks for multivariate time series is spatial-temporal graph neural networks. Spatial-temporal graph neural networks take multivariate time series and an external graph structure as inputs, and they aim to predict future values or labels of multivariate time series. Spatial-temporal graph neural networks have achieved significant improvements compared to methods that do not utilize structural information. 
However, these approaches still fall short for modeling multivariate time series due to the following challenges:
\begin{itemize}
    \item \textit{Challenge 1: Unknown Graph Structure.} Existing GNN approaches rely heavily on a pre-defined graph structure in order to perform time series forecasting. In most cases, multivariate time series does not have an explicit graph structure. The relationships among variables has to be discovered from data rather than being provided as ground truth knowledge.
    \item \textit{Challenge 2: Graph Learning \& GNN Learning. } Even though a graph structure is available, most GNN approaches focus only on message passing (GNN Learning) and overlook the fact that the graph structure is not optimal and should be updated during training. The question then is how to simultaneously learn the graph structure and the GNN for time series in an end-to-end framework.
\end{itemize}

\begin{figure}
	\centering
	\scalebox{0.5}{\includegraphics[width=\textwidth]{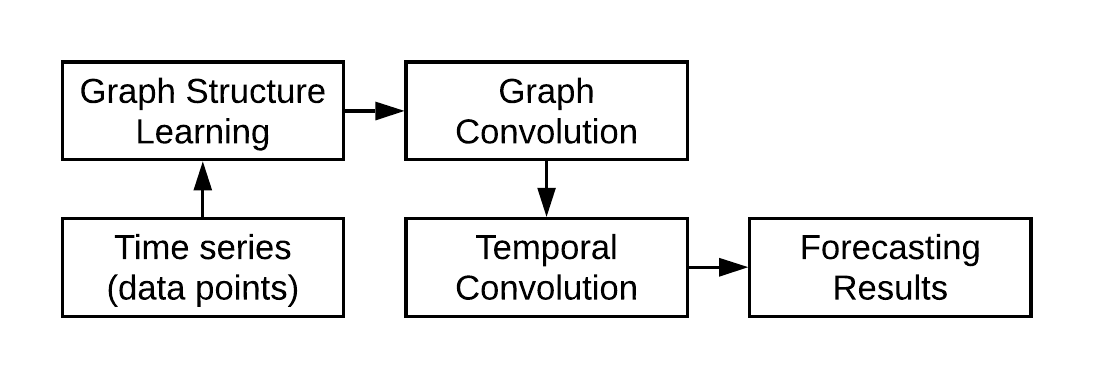}}
	\caption{A concept map of our proposed framework. }
	\label{fig:concept}
\end{figure}

In this paper, we propose a novel approach to overcome these challenges. As demonstrated by Figure \ref{fig:concept}, our framework consists of three core components - the graph learning layer, the graph convolution module, and the temporal convolution module. For \textit{Challenge 1}, we propose a novel graph learning layer, which extracts a sparse graph adjacency matrix adaptively based on data. Furthermore, we develop a graph convolution module to address the spatial dependencies among variables, given the adjacency matrix computed by the graph learning layer. This is designed specifically for directed graphs and avoids the over-smoothing problem that frequently occurs in graph convolutional networks. Finally, we propose a temporal convolution module to capture temporal patterns by modified 1D convolutions. It can both discover temporal patterns with multiple frequencies and process very long sequences.

As all parameters are learnable through gradient descent, the proposed framework is able to model multivariate time series data and learn the internal graph structure simultaneously in an end-to-end manner (for \textit{Challenge 2}). To reduce the difficulty of solving a highly non-convex optimization problem and to reduce memory occupation in processing large graphs, we propose a learning algorithm that uses a curriculum learning strategy to find a better local optimum and splits multivariate time series into subgroups during training. The advantages here are that our proposed framework is generally applicable to both \textbf{small and large graphs, short and long time series, with and without externally defined graph structures}. In summary, our main contributions are as follows:

\begin{itemize}
    \item To the best of our knowledge, this is the first study on multivariate time series data generally from a graph-based perspective with graph neural networks.
    
    \item We propose a novel graph learning module to learn hidden spatial dependencies among variables. Our method opens a new door for GNN models to handle data without explicit graph structure. 
    
    \item We present a joint framework for modeling multivariate time series data and learning graph structures. Our framework is more generic than any existing spatial-temporal graph neural network as it can handle multivariate time series with or without a pre-defined graph structure.
    
    \item Experimental results show that our method outperforms the state-of-the-art  methods on 3 of 4 benchmark datasets and achieves on-par performance with other GNNs on two traffic datasets which provide extra structural information.
\end{itemize}


\section{Backgrounds}
\subsection{Multivariate Time Series Forecasting}
Time series forecasting has been studied for a long time. The majority of existing methods follow a statistical approach. The auto-regressive integrated moving average (ARIMA) \cite{box2015time} generalizes a family of a linear model, including auto-regressive (AR), moving average (MA), and auto-regressive moving average (ARMA). The vector auto-regressive model (VAR) extends the AR model to capture the linear interdependencies among multiple time series. Similarly, the vector auto-regressive moving average model (VARMA) is proposed as a multivariate version of the ARMA model. Gaussian process (GP), as a Bayesian approach, models the distribution of a multivariate variable over functions. GP can be applied naturally to model multivariate time series data \cite{frigola2015bayesian}. Although statistical models are widely used in time series forecasting due to their simplicity and interpretability, they make strong assumptions with respect to a stationary process and they do not scale well to multivariate time series data. Deep-learning-based approaches are free from stationary assumptions and they are effective methods to capture non-linearity. Lai et al. \cite{lai2018modeling} and Shih et al. \cite{shih2019temporal} are the first two deep-learning-based models designed for multivariate time series forecasting. They employ convolutional neural networks to capture local dependencies among variables and recurrent neural networks to preserve long-term temporal dependencies.
Convolutional neural networks encapsulate interactions among variables into a global hidden state. Therefore, they cannot fully exploit latent dependencies between pairs of variables. 

\subsection{Graph Neural Networks}
Graph neural networks have enjoyed great success in handling spatial dependencies among entities in a network. 
Graph neural networks assume that the state of a node depends on the states of its neighbors. To capture this type of spatial dependency,  various kinds of graph neural networks have been developed through message passing \cite{gilmer2017neural},
information propagation \cite{klicpera2019predict},
and graph convolution \cite{kipf2017semi}. 
Sharing similar roles, they essentially capture a node's high-level representation by passing information from a node's neighbors to the node itself. Most recently, we have seen the emergence of a type of graph neural networks known as spatial-temporal graph neural networks. This form of neural networks is proposed initially to solve the problem of traffic prediction \cite{li2018diffusion,yu2018spatio,wu2019graph,zheng2020gman,chen2020multi} and skeleton-based action recognition \cite{yan2018spatial, shi2019two}
.  The inputs to spatial-temporal graph neural networks are multivariate time series with an external graph structure which describes the relationships among variables in multivariate time series. For spatial-temporal graph neural networks, spatial dependencies among nodes are captured by graph convolutions, while temporal dependencies among historical states are preserved by recurrent neural networks \cite{seo2018structured,li2018diffusion} or 1D convolutions \cite{yu2018spatio,yan2018spatial}.
Although existing spatial-temporal graph neural networks have achieved significant improvements compared to methods without using a graph structure, they are incapable of handling pure multivariate time series data effectively due to the absence of a pre-defined graph and lack of a general framework.

\section{Problem Formulation}

In this paper, we focus on the task of multivariate time series forecasting.  Let $\mathbf{z}_t\in \mathbf{R}^N$ denote the value of a multivariate variable of dimension $N$ at time step t, where $z_{t}[i]\in R$ denote the value of the $i^{th}$ variable at time step t.  Given a sequence of historical $P$ time steps of observations on a multivariate variable,  $\mathbf{X}=\{\mathbf{z}_{t_1},\mathbf{z}_{t_2}, \cdots, \mathbf{z}_{t_P}\}$, our goal is to predict the $Q$-step-away value of $\mathbf{Y}=\{\mathbf{z}_{t_{P+Q}}\}$, or a sequence of future values  $\mathbf{Y}=\{\mathbf{z}_{t_{P+1}},\mathbf{z}_{t_{P+2}}, \cdots, \mathbf{z}_{t_{P+Q}}\}$. 
More generally, the input signals can be coupled with other auxiliary features such as time of the day, day of the week, and day of the season.
Concatenating the input signals with auxiliary features, we assume the inputs instead are  $\mathbf{\mathcal{X}}=\{\mathbf{S}_{t_1},\mathbf{S}_{t_2}, \cdots, \mathbf{S}_{t_P}\}$ where $\mathbf{S}_{t_i}\in \mathbf{R}^{N\times D}$, $D$ is the feature dimension, the first column of $\mathbf{S}_{t_i}$ equals to $\mathbf{z}_{t_i}$, and the rest are auxiliary features. We aim to build a mapping $f(\cdot)$ from $\mathbf{\mathcal{X}}$ to $\mathbf{Y}$  by minimizing the absolute loss with $l2$ regularization. 

Graphs describe the relationships among entities in a network. We give a formal definition of graph-related concepts below.
	\begin{definition}[Graph]
		A graph is formulated as $G$ = $(V,E)$ where $V$ is the set of nodes, and $E$ is the set of edges. We use $N$ to denote the number of nodes in a graph.
	\end{definition}

	\begin{definition}[Node Neighborhood]
    Let $v \in V$ to denote a node and $e =(v,u) \in E$ to denote an edge pointing from $u$ to $v$. The neighborhood of a node $v$ is defined as $N(v)= \{u\in V| (v,u)\in E\}$.
	\end{definition}

    \begin{definition}[Adjacency Matrix]
		The adjacency matrix is a mathematical representation of a graph, denoted as $\mathbf{A}\in R^{N\times N}$ with $A_{ij} = c>0$ if $(v_i,v_j) \in E$ and $A_{ij} = 0$ if $(v_i,v_j) \notin E$. 
	\end{definition}

From a graph-based perspective, we consider variables in multivariate time series as nodes in graphs. We describe the relationships among nodes using the graph adjacency matrix. The graph adjacency matrix is not given by the multivariate time series data in most cases and will be learned by our model. 

\section{Framework of MTGNN}\label{sec:met}

\begin{figure*}
	\centering
	\scalebox{.8}{\includegraphics[width=\textwidth]{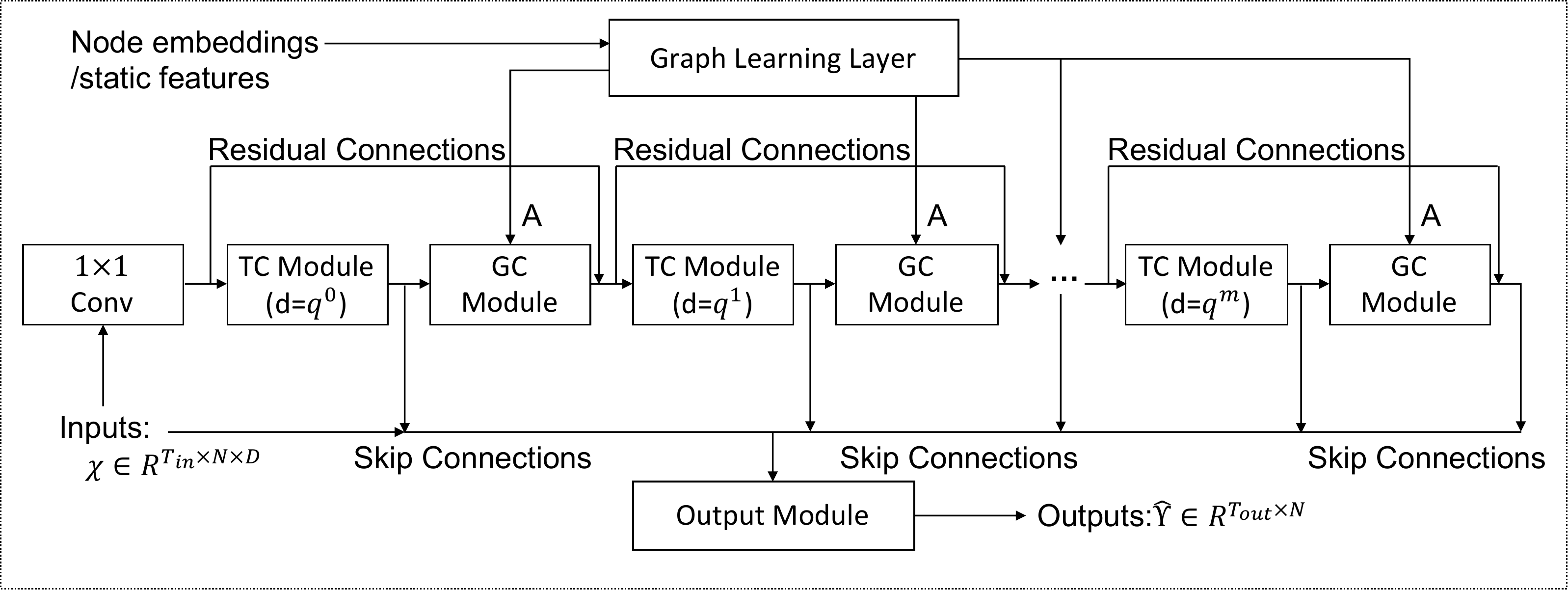}}
	\caption{The framework of MTGNN. A $1\times 1$ standard convolution first projects the inputs into a latent space. Afterward, temporal convolution modules and graph convolution modules are interleaved with each other to capture temporal and spatial dependencies respectively. The hyper-parameter, dilation factor $d$, which controls the receptive field size of a temporal convolution module, is increased at an exponential rate of $q$. The graph learning layer learns the hidden graph adjacency matrix, which is used by graph convolution modules. Residual connections and skip connections are added to the model to avoid the problem of gradient vanishing. The output module projects hidden features to the desired dimension to get the final results.}
	\label{fig:model}
\end{figure*}

\begin{figure}
	\centering
	\scalebox{0.4}{\includegraphics[width=\textwidth]{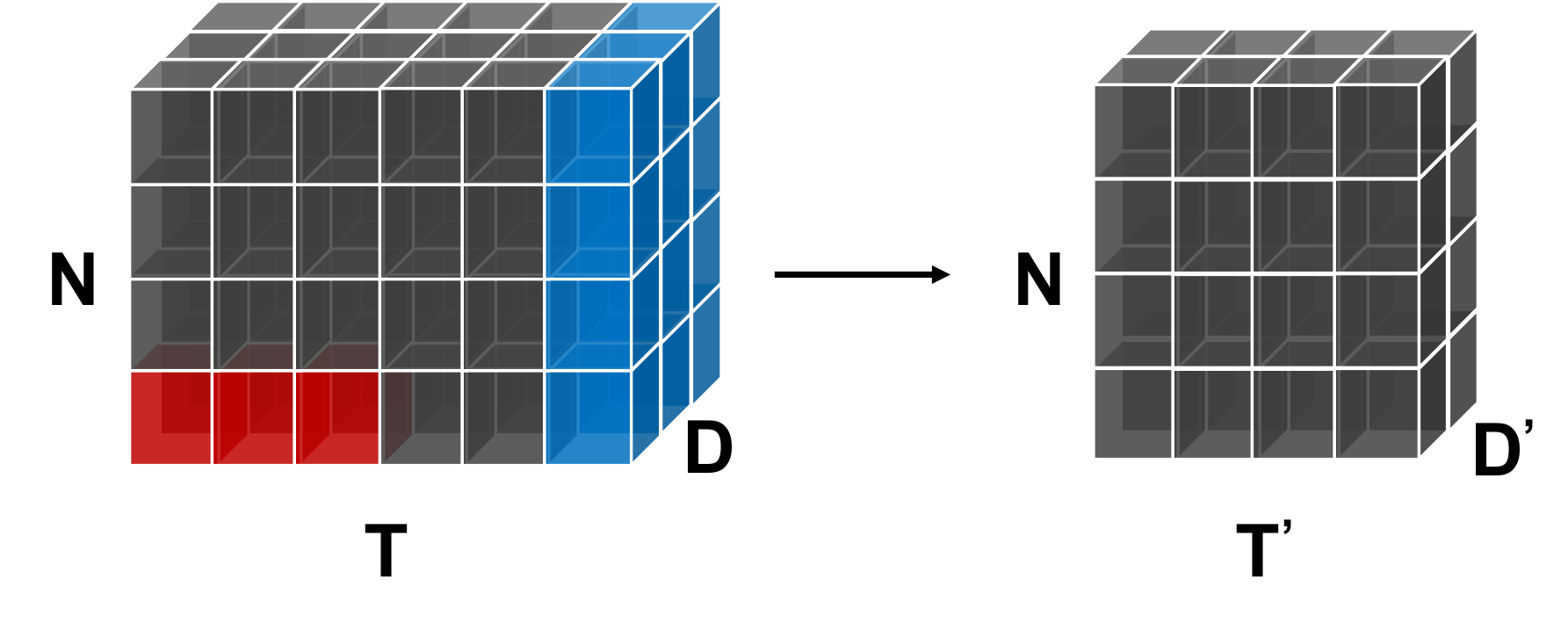}}
	\caption{\small A demonstration of how a temporal convolution module and a graph convolution module collaborate with each other. A temporal convolution module filters the inputs by sliding a 1D window over the time and node axes, as denoted by the red. A graph convolution module filters the inputs at each step, denoted by the blue.}
	\label{fig:cube}
\end{figure}

\subsection{Model Architecture}
We first elaborate on the general framework of our model. As illustrated in Figure \ref{fig:model}, MTGNN on the highest level consists of a \textit{graph learning layer}, $m$ \textit{graph convolution modules}, $m$ \textit{temporal convolution modules}, and an output module. To discover hidden associations among nodes, a graph learning layer computes a graph adjacency matrix, which is later used as an input to all graph convolution modules. Graph convolution modules are interleaved with temporal convolution modules to capture spatial and temporal dependencies respectively. Figure \ref{fig:cube} gives a demonstration of how a temporal convolution module and a graph convolution module collaborate with each other. To avoid the problem of gradient vanishing, residual connections are added from the inputs of a temporal convolution module to the outputs of a graph convolution module. Skip connections are added after each temporal convolution module. To get the final outputs,  the output module projects the hidden features to the desired output dimension. In more detail, the core components of our model are illustrated in the following:

\subsection{Graph Learning Layer}
The graph learning layer learns a graph adjacency matrix adaptively to capture the hidden relationships among time series data.  To construct a graph, existing studies measure the similarity between pairs of nodes by a distance metric, such as dot product and Euclidean distance \cite{li2018diffusion}.  This leads inevitably to the problem of high time and space complexity with $O(N^2)$. It means the computation and memory cost grows quadratically with the increase of graph size. This restricts the model's capability of handling larger graphs. To address this limitation, we adopt a sampling approach, which only calculates pair-wise relationships among a subset of nodes. This cuts off the bottleneck of computation and memory in each minibatch. More details will be provided in Section \ref{sec:algo}. 

Another problem is that existing distance metrics are often symmetric or bi-directional. In multivariate time series forecasting, we expect that the change of a node's condition causes the change of another node's condition such as traffic flow. Therefore the learned relation is supposed to be uni-directional. Our proposed graph learning layer is specifically designed to extract uni-directional relationships, illustrated as follows:
\begin{align}
    \mathbf{M}_1 &= tanh(\alpha \mathbf{E}_1\mathbf{\Theta}_1)  \label{eq:m1}\\
    \mathbf{M}_2 &= tanh(\alpha \mathbf{E}_2\mathbf{\Theta}_2)
    \label{eq:m2}\\
    \mathbf{A} &= ReLU(tanh(\alpha (\mathbf{M}_1\mathbf{M}_2^T-\mathbf{M}_2\mathbf{M}_1^T))) \label{eq:adp}\\ 
    for&\;i=1,2,\cdots,N \\
    &\mathbf{idx}= argtopk(\mathbf{A}[i,:]) \label{eq:argtop}\\
    &\mathbf{A}[i,-\mathbf{idx}] = 0, \label{eq:adpo}
\end{align}
where $\mathbf{E}_1,\mathbf{E}_2$ represents randomly initialized node embeddings, which are learnable during training, $\Theta_1,\Theta_2$ are model parameters, $\alpha$ is a hyper-parameter for controlling the saturation rate of the activation function, and $argtopk(\cdot)$ returns the index of the top-k largest values of a vector. The asymmetric property of our proposed graph adjacency matrix is achieved by Equation \ref{eq:adp}. The subtraction term and the ReLU activation function regularize the adjacency matrix so that if $A_{vu}$ is positive, its diagonal counterpart $A_{uv}$ will be zero.  
Equation \ref{eq:argtop}-\ref{eq:adpo} is a strategy to make the adjacency matrix sparse while reducing the computation cost of the following graph convolution. For each node, we select its top-k closest nodes as its neighbors. While retaining the weights for connected nodes, we set the weights of non-connected nodes as zero.

\textit{Incorporate External Data.} The inputs to the graph learning layer are not limited to node embeddings. In case that external knowledge about the attributes of each node is given, we can also set $\mathbf{E}_1=\mathbf{E}_2=\mathbf{Z}$, where $\mathbf{Z}$ is a static node feature matrix. Some works have considered capturing dynamic spatial dependencies \cite{guo2019attention,shi2019two}. In other words, they dynamically adjust the weight of two connected nodes based on temporal inputs. However, assuming dynamic spatial dependencies makes the model extremely hard to converge when we need to learn the graph structure at the same time. The advantage of our approach is that we can learn stable and interpretable node relationships over the period of the training dataset. Once the model is trained in an on-line learning version, our graph adjacency matrix is also adaptable to change as new training data updates the model parameters.

\subsection{Graph Convolution Module}
\begin{figure}
        \centering
        \begin{subfigure}[b]{0.23\textwidth}
            \centering
            \includegraphics[width=\textwidth]{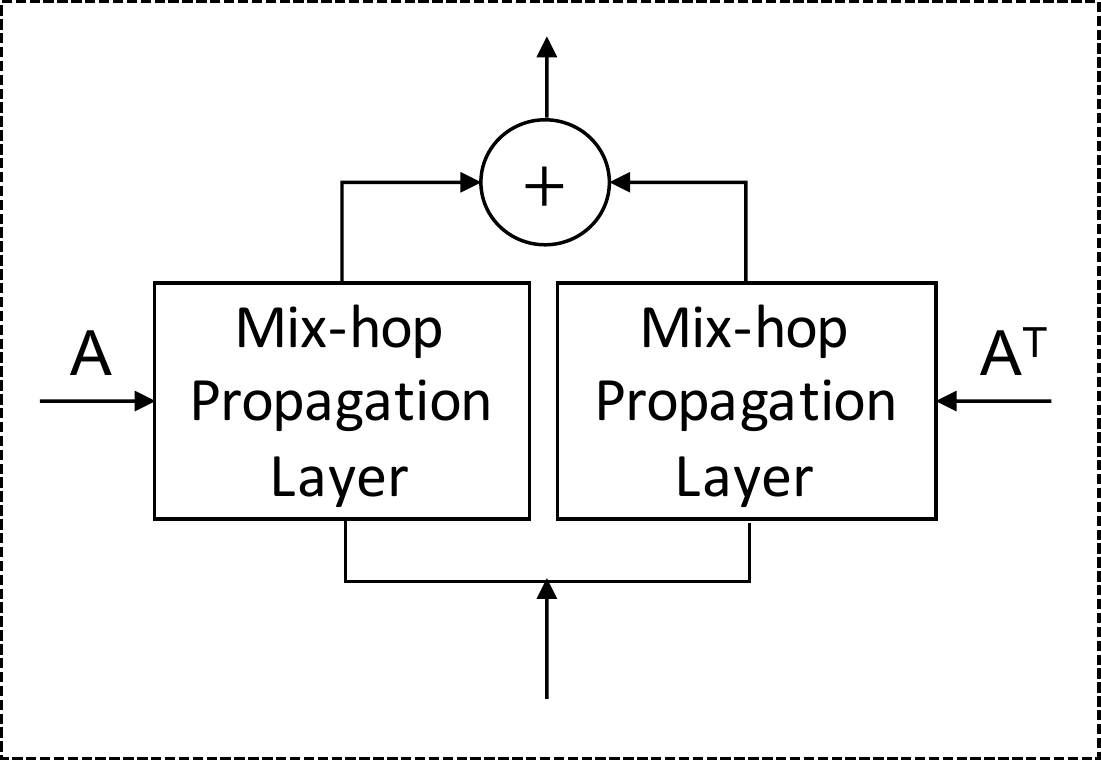}
            \caption[GC module]%
            {{\small GC module}}    
            \label{fig:gcm}
        \end{subfigure}
        \hfill
        \begin{subfigure}[b]{0.23\textwidth}  
            \centering 
            \includegraphics[width=\textwidth]{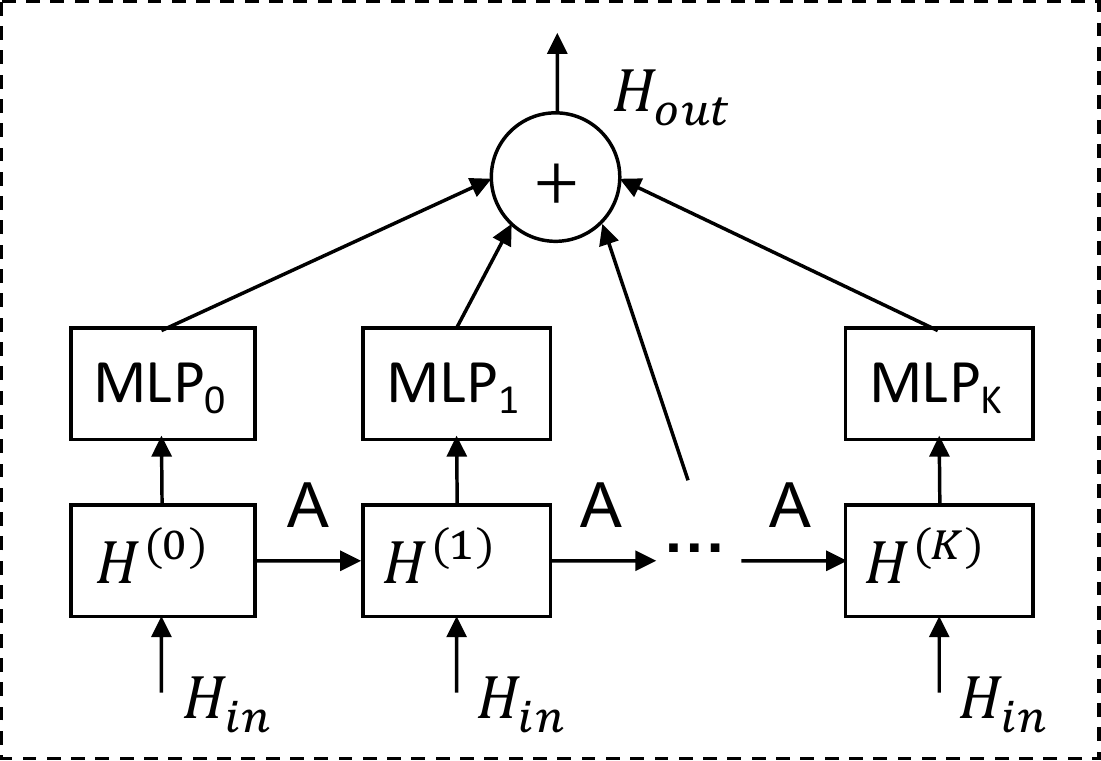}
            \caption[]%
            {{\small Mix-hop propagation layer}}    
            \label{fig:mixprop}
        \end{subfigure}
        \caption[Graph convolution module and mix-hop propagation layer]
        {\small Graph convolution and mix-hop propagation layer.} 
        \label{fig:gc}
\end{figure}

The graph convolution module aims to fuse a node's information with its neighbors' information to handle spatial dependencies in a graph. The graph convolution module consists of two mix-hop propagation layers to process inflow and outflow information passed through each node separately. The net inflow information is obtained by adding the outputs of the two mix-hop propagation layers.  Figure \ref{fig:gc} shows the architecture of the graph convolution module and the mix-hop propagation layer. 

\paragraph{Mix-hop Propagation Layer.}
Given a graph adjacency matrix, we propose the mix-hop propagation layer to handle information flow over spatially dependent nodes. The proposed mix-hop propagation layer consists of two steps - the information propagation step and the information selection step. We first give the mathematical form of these two steps and then illustrate our motivations. The information propagation step is defined as follows:
\begin{equation}
\mathbf{H}^{(k)}=\beta\mathbf{H}_{in}+ (1-\beta)\Tilde{\mathbf{A}}\mathbf{H}^{(k-1)},
\label{eq:ppnp}
\end{equation}
where $\beta$ is a hyper parameter, which controls the ratio of retaining the root node's original states. The information selection step is defined as follows
\begin{equation}
\mathbf{H}_{out} = \sum_{i=0}^{K}\mathbf{H}^{(k)}\mathbf{W}^{(k)},
\label{eq:mix}
\end{equation}
where $K$ is the depth of propagation, $\mathbf{H}_{in}$ represents the input hidden states outputted by the previous layer, $\mathbf{H}_{out}$ represents the output hidden states of the current layer, $\mathbf{H}^{(0)}=\mathbf{H}_{in}$, $\Tilde{\mathbf{A}}=\Tilde{\mathbf{D}}^{-1}(\mathbf{A}+\mathbf{I})$, and $\Tilde{\mathbf{D}}_{ii}=1+\sum_j\mathbf{A}_{ij}$.
In Figure \ref{fig:mixprop}, we demonstrate the information propagation step and information selection step in the proposed mix-hop propagation layer. It first propagates information horizontally and selects information vertically.

The information propagation step propagates node information along with the given graph structure recursively.  A severe limitation of graph convolutional networks is that node hidden states converge to a single point as the number of graph convolution layers goes to infinity. This is because the graph convolutional network with many layers reaches the random walk's limit distribution regardless of the initial node states.  To address this problem,  motivated by Klicpera et al. \cite{klicpera2019predict}, we retain a proportion of nodes' original states during the propagation process so that the propagated node states can both preserve locality and explore a deep neighborhood. However, if we only apply Equation \ref{eq:ppnp}, some node information will be lost. Under the extreme circumstance that no spatial dependencies exist, aggregating neighborhood information simply adds useless noises to each node. Therefore, the information selection step is introduced to filter out important information produced at each hop. According to Equation \ref{eq:mix}, the parameter matrix $\mathbf{W}^{(k)}$ functions as a feature selector. When the given graph structure does not entail spatial dependencies, Equation \ref{eq:mix} is still able to preserve the original node-self information by adjusting $\mathbf{W}^{(k)}$ to 0 for all $k>0$.

\textit{Connection to existing works.} The idea of mix-hop has been explored by \cite{kapoor2019mixhop} and \cite{chen2019dagcn}. Kapoor et al. \cite{kapoor2019mixhop} concatenate information from different hops. Chen et al. \cite{chen2019dagcn} propose an attention mechanism to weight information among different hops. They both apply GCN for information propagation. However, as GCN faces the over-smoothing problem, information from higher hops may not or negatively contribute to the overall performance.  To avoid this, our approach keeps a balance between local and neighborhood information.  Furthermore, Kapoor et al. \cite{kapoor2019mixhop} show that their proposed model with two mix-hop layers has the capability to represent the delta difference between two consecutive hops. Our approach can achieve the same effect with only one mix-hop propagation layer. Suppose $K=2$, $\mathbf{W}^{(0)}=\mathbf{0}$, $\mathbf{W}^{(1)}=-\mathbf{1}$, and $\mathbf{W}^{(2)}=\mathbf{1}$,  then
\begin{equation}
    \mathbf{H}_{out}=\Delta(\mathbf{H}^{(2)},\mathbf{H}^{(1)})= \mathbf{H}^{2}-\mathbf{H}^{1}.
\end{equation}
From this perspective, using summation is more efficient to represent all linear interactions of different hops compared with the concatenation method.

\begin{figure}
        \begin{subfigure}[b]{0.23\textwidth}   
            \centering 
            \includegraphics[width=\textwidth]{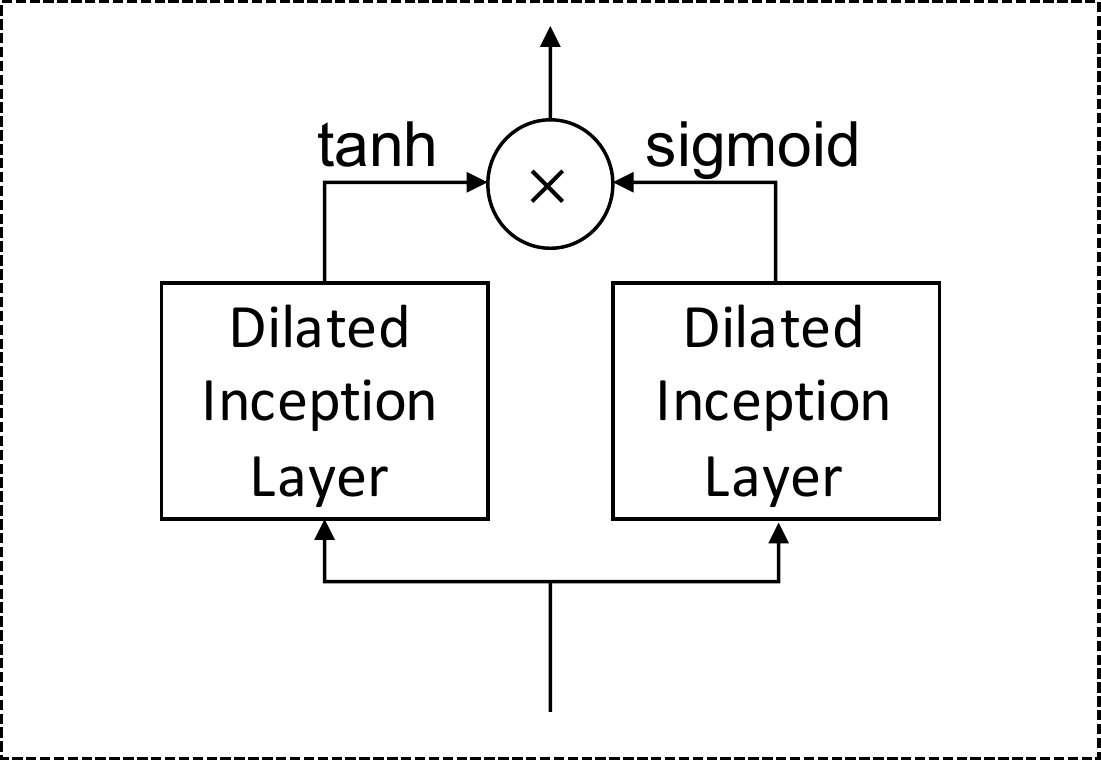}
            \caption[TC module]%
            {{\small TC module}}    
            \label{fig:tcm}
        \end{subfigure}
        \hfill
        \begin{subfigure}[b]{0.23\textwidth}   
            \centering 
            \includegraphics[width=\textwidth]{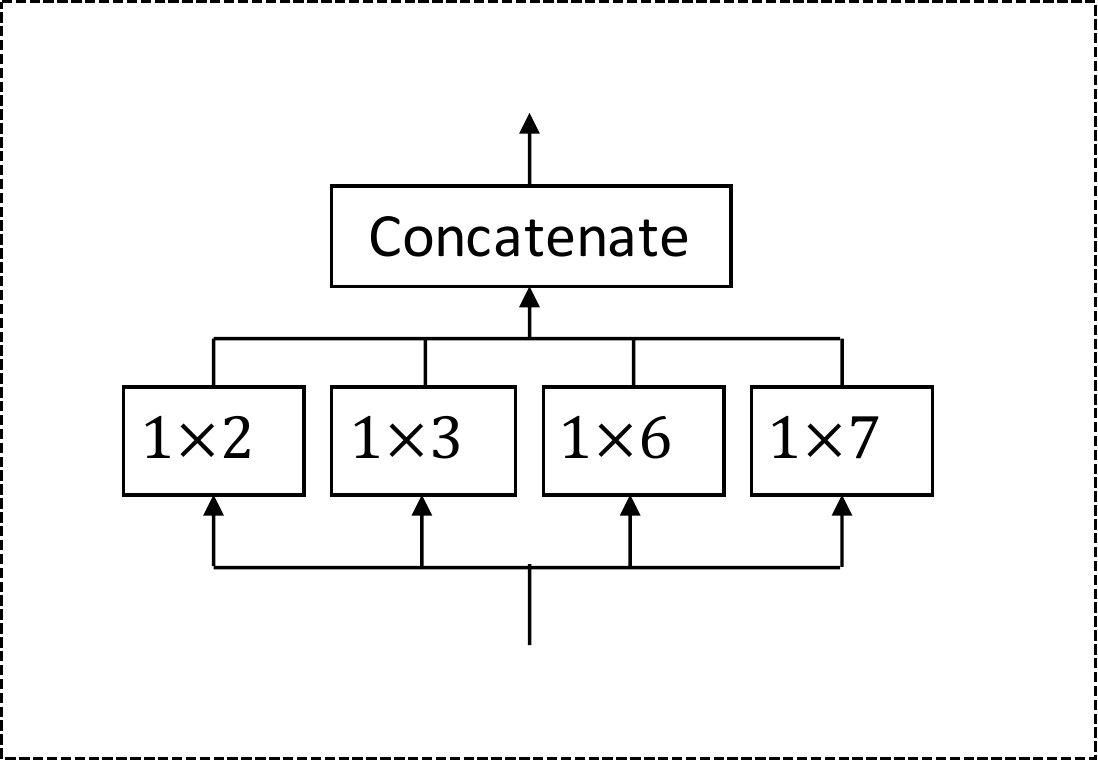}
            \caption[Dilated inception layer]%
            {{\small Dilated inception layer}}    
            \label{fig:dil}
        \end{subfigure}
        \caption[Temporal convolution module and dilated inception layer.]
        {\small The temporal convolution and dilated inception layer.} 
        \label{fig:tc}
\end{figure}
\subsection{Temporal Convolution  Module}
The temporal convolution module applies a set of standard dilated 1D convolution filters to extract high-level temporal features. This module consists of two dilated inception layers. One dilated inception layer is  followed by a tangent hyperbolic activation function and works as a filter. The other layer is followed by a sigmoid activation function and functions as a gate to control the amount of information that the filter can pass to the next module. Figure \ref{fig:tc} shows the architecture of the temporal convolution module and the dilated inception layer. 

\paragraph{Dilated Inception Layer}
The temporal convolution module captures sequential patterns of time series data through 1D convolutional filters. To come up with a temporal convolution module that is able to both discover temporal patterns with various ranges and handle very long sequences, we propose the dilated inception layer which combines two widely applied strategies from convolutional neural networks, i.e., using filters with multiple sizes \cite{szegedy2015going} and applying dilated convolution \cite{yu2016multi}.

First, choosing the right kernel size is a challenging problem for convolutional networks. The filter size can be too large to represent short-term signal patterns subtly, or too small to discover long-term signal patterns sufficiently.
In image processing, a widely employed strategy is called inception, which concatenates the outputs of 2D convolution filters with three different kernel sizes, $1\times1$, $3\times3$, and $5\times 5$. Moving from 2D images to 1D time series, the set of $1\times1$, $1\times3$, and $1\times 5$ filter sizes do not suit the nature of temporal signals. As temporal signals tend to have several inherent periods such as $7$, $12$, $24$, $28$, and $60$, a stack of inception layers with filter size $1\times1$, $1\times3$, and $1\times 5$ cannot well encompass those periods. Alternatively, we propose a temporal inception layer consisting of four filter sizes, viz. $1\times2$, $1\times3$, $1\times6$, and $1\times7$. The aforementioned periods can all be covered by the combination of these filter sizes. For example, to represent the period $12$, a model can pass the inputs through a $1\times 7$ filter from the first temporal inception layer followed by a $1\times 6$ filter from the second temporal inception layer. 

Second, the receptive field size of a convolutional network grows in a linear progression with the depth of the network and the kernel size of the filter. Consider a convolutional network with $m$ $1D$ convolution layers of kernel size $c$, the receptive field size of the convolutional network is, 
\begin{equation}
R = m(c-1)+1.
\end{equation}
To process very long sequences, it requires either a very deep network or very large filters. We adopt dilated convolution to reduce model complexity. Dilated convolution operates a standard convolution filter on down-sampled inputs with a certain frequency.  For example, where the dilation factor is 2, it applies standard convolution on inputs sampled every two steps. 
Following \cite{oord2016wavenet}, we let the dilation factor for each layer increase exponentially at a rate of $q\;\;(q>1)$. Suppose the initial dilation factor is $1$, the receptive field size of a $m$ layer dilated convolutional network with kernel size $c$ is 
\begin{equation}
R = 1 + (c-1)(q^m-1)/(q-1).
\end{equation}
This indicates that the receptive field size of the network also grows exponentially with an increase in the number of hidden layers at the rate of $q$. Therefore, using this dilation strategy can capture much longer sequences than proceeding without it.

Formally, combining inception and dilation, we propose the dilated inception layer, demonstrated by Figure \ref{fig:dil}. Given a 1D sequence input $\mathbf{z}\in \mathbf{R}^T$ and filters consisting of $\mathbf{f}_{1\times2} \in \mathbf{R}^2$, $\mathbf{f}_{1\times3} \in \mathbf{R}^3$, $\mathbf{f}_{1\times6} \in \mathbf{R}^6$, and $\mathbf{f}_{1\times7} \in \mathbf{R}^7$, our dilated inception layer takes the form,

\begin{equation}
    \mathbf{z} =  concat(\mathbf{z}\star \mathbf{f}_{1\times2},\mathbf{z}\star \mathbf{f}_{1\times3},\mathbf{z}\star \mathbf{f}_{1\times6},\mathbf{z}\star \mathbf{f}_{1\times7}),
\end{equation}
where the outputs of the four filters are truncated to the same length according to the largest filter and concatenated across the channel dimension, and the dilated convolution denoted by $\mathbf{z}\star \mathbf{f}_{1\times k}$ is defined as
\begin{equation}
    \mathbf{z}\star \mathbf{f}_{1\times k}(t) = \sum_{s=0}^{k-1} \mathbf{f}_{1\times k}(s)\mathbf{z}(t-d\times s),
\end{equation}
where $d$ is the dilation factor.

\subsection{Skip Connection Layer \& Output Module}
Skip connection layers are essentially $1\times L_i$ standard convolutions where $L_i$ is the sequence length of the inputs to the $i^{th}$ skip connection layer. It standardizes information that jumps to the output module to have the same sequence length $1$. The output module consists of two $1\times1$ standard convolution layers, transforming the channel dimension of the inputs to the desired output dimension. In case we want to predict a certain future step only, the desired output dimension is $1$. When we want to predict $Q$ consecutive steps, the desired output dimension is $Q$.

\subsection{Proposed Learning Algorithm}
\label{sec:algo}
We propose a learning algorithm to enhance our model's capability of handling large graphs and stabilizing in a better local optimum. Training on a graph often requires storing all node intermediate states into memory. If a graph is large, it will face the problem of memory overflow. Most relevant to us, Chiang et al.  \cite{chiang2019cluster} propose a sub-graph training algorithm to tackle the memory bottleneck. They apply a graph clustering algorithm to partition a graph into sub-graphs and train a graph convolutional network on the partitioned sub-graphs. In our problem, it is not practical to cluster nodes based on their topological information because our model learns the latent graph structure at the same time. Alternatively, in each iteration, we randomly split the nodes into several groups and let the algorithm learn a sub-graph structure based on the sampled nodes. This gives each node the full possibilities of being assigned with another node in one group so that the similarity score between these two nodes can be computed and updated. As a side benefit, if we split the nodes into $s$ groups, we can reduce the time and space complexity of our graph learning layer from $O(N^2)$ to $(N/s)^2$ in each iteration. After training, as all node embeddings are well-trained, a global graph can be constructed to fully utilize spatial relationships. Although it is computationally expensive, the adjacency matrix can be pre-computed in parallel before making predictions. 

The second consideration of our proposed algorithm is to facilitate our model stabilize in a better local optimum. In the task of multi-step forecasting, we observe that long-term predictions often achieve greater improvements than those in the short-term in terms of model performance. We believe the reason is that our model predicts multi-steps altogether, and long-term predictions produce a much higher loss than short-term predictions. As a result, to minimize the overall loss, the model focuses more on improving the accuracy of long-term predictions. To address this issue we propose a \textit{curriculum learning} strategy for the multi-step forecasting task. The algorithm starts with solving the easiest problem, predicting the next one-step only. It is very advantageous for the model to find a good starting point. With the increase in iteration numbers, we increase the prediction length of the model gradually so that the model can learn the hard task step by step.
Covering all this, our algorithm is given in Algorithm 1.  Further complexity analysis of our model can be found in Appendix \textbf{A.1}.

\begin{algorithm}[t]
\caption{The learning algorithm of MTGNN.}\label{alg}
\begin{algorithmic}[1]
\small
\State \textbf{Input}: The dataset $O$, node set $V$, the initialized MTGNN model $f(\cdot)$ with $\Theta$, learning rate $\gamma$, batch size $b$, step size $s$, split size $m$ (default=1).
\State set $iter=1,r=1$
\Repeat
\State sample a batch $(\mathcal{X}\in R^{b\times T\times N \times D}, \mathcal{Y} \in R^{b\times T'\times N} )$ from $O$.
\State random split the node set $V$ into $m$ groups, $\cup_{i=1}^m V_i=V$.
\If{$iter\%s==0$ and $r<=T'$}
\State $r=r+1$
\EndIf

\For{i in 1:m}
\State compute $\hat{\mathcal{Y}} = f(\mathcal{X}[: , : , id(V_i) , :];\mathbf{\Theta})$
\State compute  $L=loss(\hat{\mathcal{Y}}[: , :r , :],\mathcal{Y}[: , :r , id(V_i)])$
\State compute the stochastic gradient of $\Theta$ according to $L$.
\State update model parameters $\Theta$ according to their gradients and the learning rate $\gamma$.
\EndFor
\State $iter=iter+1$.
\Until{convergence}
\end{algorithmic}
\end{algorithm}

\section{Experimental Studies}
We validate MTGNN on two tasks - both single-step and multi-step forecasting. First, we compare the performance of MTGNN with other multivariate time series models on four benchmark datasets for multivariate time series forecasting, where the aim is to predict a single future step.
Furthermore, to show how well MTGNN performs, compared with other spatial-temporal graph neural networks which, in contrast, use pre-defined graph structural information, we evaluate MTGNN on two benchmark datasets for spatial-temporal graph neural networks, where the aim is to predict multiple future steps. Further results on parameter study can be found in Appendix \textbf{A.4}.

\subsection{Experimental Setting}
In Table \ref{tb:data-stats}, we summarize statistics of benchmark datasets.
More details about the datasets is given in Appendix \textbf{A.2}. 
We use five evaluation metrics, including Mean Absolute Error (MAE), Root Mean Squared Error (RMSE), Mean Absolute Percentage Error (MAPE), Root Relative Squared Error (RRSE), and Empirical Correlation Coefficient (CORR). For RMSE, MAE, MAPE, and RRSE, lower values are better. For CORR, higher values are better. 
Other experimental setups are given in Appendix \textbf{A.3}.
\begin{table}
	\begin{center}
		\caption{\small Dataset statistics.}
		\label{tb:data-stats}
		\resizebox{\columnwidth}{!}{%
			\begin{tabular}{lrrrrrr} 
				\toprule
				Datasets 		& \# Samples		& \# Nodes	& Sample Rate	& Input Length & Output Length	 \\
				\midrule
				traffic 		& 17,544 	& 862 	& 1 hour & 168 & 1	\\
				solar-energy			& 52,560	& 137	& 10 minutes & 168 & 1\\
				electricity 	& 26,304 	& 321 	& 1 hour & 168 & 1	\\
				exchange-rate		& 7,588		& 8		& 1 day	& 168 & 1	\\
				\midrule
				metr-la & 34272 & 207 & 5 minutes & 12 & 12 \\
				pems-bay & 52116 & 325 & 5 minutes & 12 & 12 \\
				\bottomrule
		\end{tabular}}
	\end{center}
\end{table}

\subsection{Baseline Methods for Comparision}
MTGNN and MTGNN+sampling are our models to be evaluated. MTGNN is our proposed model.   MTGNN+sampling is our proposed model trained on a sampled subset of a graph in each iteration. 
Baseline methods are summarized in the following:

\subsubsection{Single-step forecasting}
\begin{itemize}
	\item AR: An auto-regressive model.
	\item VAR-MLP: A hybrid model of the multilayer perception (MLP) and auto-regressive model (VAR) \cite{zhang2003time}.
	\item GP: A Gaussian Process time series model \cite{roberts2013gaussian,frigola2016bayesian}.
	\item RNN-GRU: A recurrent neural network with fully connected GRU hidden units.
	\item LSTNet: A deep neural network, which combines convolutional neural networks and recurrent neural networks  \cite{lai2018modeling}.
	\item TPA-LSTM: An attention-recurrent neural network \cite{shih2019temporal}.
\end{itemize}
\subsubsection{Multi-step forecasting}
\begin{itemize}
	\item DCRNN: A diffusion convolutional recurrent neural network, which combines diffusion graph convolutions with recurrent neural networks \cite{li2018diffusion}.
	\item STGCN: A spatial-temporal graph convolutional network,  which incorporates graph convolutions with 1D convolutions \cite{yu2018spatio}.
	\item Graph WaveNet: A spatial-temporal graph convolutional network, which integrates diffusion graph convolutions with 1D dilated convolutions \cite{wu2019graph}.
	\item ST-MetaNet: A sequence-to-sequence architecture, which employs meta networks to generate parameters \cite{pan2019urban}.
	\item GMAN: A graph multi-attention network with spatial and temporal attentions \cite{zheng2020gman}. 
	\item MRA-BGCN: A multi-range attentive bicomponent GCN \cite{chen2020multi}.
\end{itemize}

\subsection{Main Results}

\begin{table*}
	\centering
	\caption{Baseline comparison under single-step forecasting for multivariate time series methods.}
	\label{table:single}
	\resizebox{\textwidth}{!}{
		
		\begin{tabular}{ll|cccc|cccc|cccc|cccc}

			\toprule
			\\[-1em]
			\\[-1em]
			\\[-1em]
			\multicolumn{2}{c|}{Dataset} & \multicolumn{4}{|c|}{Solar-Energy} & \multicolumn{4}{|c|}{Traffic} & \multicolumn{4}{|c|}{Electricity} & \multicolumn{4}{|c}{Exchange-Rate}
			\\
			\\[-1em]
			\\[-1em]
			\\[-1em]
			\midrule
			\\[-1em]
			\\[-1em]
			\\[-1em]
			\multicolumn{2}{c|}{} & \multicolumn{4}{|c|}{Horizon} & \multicolumn{4}{|c|}{Horizon} & \multicolumn{4}{|c|}{Horizon} & \multicolumn{4}{|c}{Horizon}  \\
			\\[-1em]
			\\[-1em]
			\\[-1em]
			\hline
			\\[-1em]
			\\[-1em]
			\\[-1em]
			
			Methods & Metrics & 3 & 6 & 12 & 24 & 3 & 6 & 12 & 24 & 3 & 6 & 12 & 24 & 3 & 6 & 12 & 24 \\
			\\[-1em]
			\\[-1em]
			\\[-1em]
			\hline
			\\[-1em]
			\\[-1em]
			\\[-1em]
			AR&RSE&0.2435&0.3790&0.5911&0.8699&0.5991&0.6218&0.6252&0.63&0.0995&0.1035&0.1050&0.1054&0.0228&0.0279&0.0353&0.0445\\
			&CORR&0.9710&0.9263&0.8107&0.5314&0.7752&0.7568&0.7544&0.7519&0.8845&0.8632&0.8591&0.8595&0.9734&0.9656&0.9526&0.9357\\
			\\[-1em]
			\\[-1em]
			\\[-1em]
			\hline
			\\[-1em]
			\\[-1em]
			\\[-1em]
			VARMLP&RSE&0.1922&0.2679&0.4244&0.6841&0.5582&0.6579&0.6023&0.6146&0.1393&0.1620&0.1557&0.1274&0.0265&0.0394&0.0407&0.0578\\
			&CORR&0.9829&0.9655&0.9058&0.7149&0.8245&0.7695&0.7929&0.7891&0.8708&0.8389&0.8192&0.8679&0.8609&0.8725&0.8280&0.7675\\
			
			\\[-1em]
			\\[-1em]
			\\[-1em]
			\hline
			\\[-1em]
			\\[-1em]
			\\[-1em]
			
			GP&RSE&0.2259&0.3286&0.5200&0.7973&0.6082&0.6772&0.6406&0.5995&0.1500&0.1907&0.1621&0.1273&0.0239&0.0272&0.0394&0.0580\\
			
			&CORR&0.9751&0.9448&0.8518&0.5971&0.7831&0.7406&0.7671&0.7909&0.8670&0.8334&0.8394&0.8818&0.8713&0.8193&0.8484&0.8278\\
			\\[-1em]
			\\[-1em]
			\\[-1em]
			\hline
			\\[-1em]
			\\[-1em]
			\\[-1em]

			RNN-GRU&RSE&0.1932&0.2628&0.4163&0.4852&0.5358&0.5522&0.5562&0.5633&0.1102&0.1144&0.1183&0.1295&0.0192&0.0264&0.0408&0.0626\\
			&CORR&0.9823&0.9675&0.9150&0.8823&0.8511&0.8405&0.8345&0.8300&0.8597&0.8623&0.8472&0.8651&0.9786&\textbf{0.9712}&0.9531&0.9223\\
			\\[-1em]
			\\[-1em]
			\\[-1em]
			\hline
			\\[-1em]
			\\[-1em]
			\\[-1em]
			LSTNet-skip	& RSE 
			& 0.1843 & 0.2559 & 0.3254 & 0.4643
			& 0.4777 & 0.4893 & 0.4950 & 0.4973
			& 0.0864 & 0.0931 & 0.1007 & 0.1007
			& 0.0226 & 0.0280 & 0.0356 & 0.0449\\
			& CORR 
			& 0.9843 & 0.9690 & 0.9467 & 0.8870 & 0.8721 & 0.8690 & 0.8614 & 0.8588
			& 0.9283 & 0.9135 & 0.9077 & 0.9119
			& 0.9735 & 0.9658 & 0.9511 & 0.9354\\ 
			\\[-1em]
			\\[-1em]
			\\[-1em]
			\hline
			\\[-1em]
			\\[-1em]
			\\[-1em]
			TPA-LSTM	& RSE 
			& 0.1803 & \textbf{0.2347} & 0.3234 & 0.4389
			& 0.4487 & 0.4658 & 0.4641 & 0.4765
			& 0.0823 & 0.0916 & 0.0964 & 0.1006
			& \textbf{0.0174} & \textbf{0.0241} & \textbf{0.0341} & \textbf{0.0444}\\
			& CORR 
			& 0.9850 & \textbf{0.9742} & 0.9487 & 0.9081 & 0.8812 & 0.8717 & 0.8717 & 0.8629
			& 0.9439 & 0.9337 & 0.9250 & 0.9133
			& \textbf{0.9790} & 0.9709 & 0.9564 & 0.9381\\ 
			\\[-1em]
			\\[-1em]
			\\[-1em]
			\hline\hline
			\\[-1em]
			\\[-1em]
			\\[-1em]
			
			MTGNN &RSE & \textbf{0.1778} & 0.2348 & \textbf{0.3109} & \textbf{0.4270}  & \textbf{0.4162}  & 0.4754 & \textbf{0.4461} & \textbf{0.4535} & \textbf{0.0745} & 0.0878 & \textbf{0.0916} & \textbf{0.0953} & 0.0194 & 0.0259 & 0.0349 & 0.0456 \\
			& CORR & \textbf{0.9852} & 0.9726 & \textbf{0.9509} & 0.9031  & \textbf{0.8963} & 0.8667 & \textbf{0.8794} & \textbf{0.8810} & \textbf{0.9474} & 0.9316 & \textbf{0.9278} & \textbf{0.9234} & 0.9786 & 0.9708 & 0.9551 & 0.9372 \\
			
			\\[-1em]
			\\[-1em]
			\\[-1em]
			\hline
			\\[-1em]
			\\[-1em]
			\\[-1em]
			MTGNN+sampling &RSE & 0.1875 & 0.2521 & 0.3347 & 0.4386 &0.4170& \textbf{0.4435}& 0.4469&0.4537&0.0762& \textbf{0.0862} &0.0938&0.0976&0.0212&0.0271&0.0350&0.0454\\
			&CORR & 0.9834  & 0.9687  & 0.9440 &  0.8990 & 0.8960& \textbf{0.8815} & 0.8793&0.8758&0.9467& \textbf{0.9354} &0.9261&0.9219&0.9788&0.9704&\textbf{0.9574}&\textbf{0.9382}\\ 
			\\[-1em]
			\\[-1em]
			\\[-1em]
			\bottomrule
		\end{tabular}
	}
\end{table*}

Table \ref{table:single} and Table \ref{table:multi} provide the main experimental results of MTGNN and MTGNN+sampling. We observe that MTGNN achieves state-of-the-art results on most of the tasks, and the performance of MTGNN only degrades marginal when it samples sub-graphs for training. In the following, we discuss experimental results of single-step and multi-step forecasting respectively.

\subsubsection{Single-step forecasting} In this experiment, we compare MTGNN with other multivariate time series models.  Table \ref{table:single} shows the experimental results for the single-step forecasting task. In general, our MTGNN achieves state-of-the-art results over almost all horizons on Solar-Energy, Traffic, and Electricity data. In particular, on Traffic data, the improvement of MTGNN in terms of RSE is significant. MTGNN lowers down RSE by 7.24\%, 3.88\%, 4.83\%  over the horizons of 3, 12, 24 on the traffic data. The main reason why MTGNN improves the results of traffic data evidently is that the nature of traffic data is better suited for our model assumption about the spatial-temporal dependencies. Obviously, the future traffic occupancy rate of a road not only depends on its past but also on its connected roads' occupancy rates. MTGNN fails to make improvements on the exchange-rate data, possibly due to the smaller graph size and fewer training examples of exchange-rate data.

\subsubsection{Multi-step forecasting}
In this experiment, we compare MTGNN with other spatial-temporal graph neural network models. Table \ref{table:multi} shows the experimental results for the task of multi-step forecasting. The significance of MTGNN lies in that it achieves on-par performance with state-of-the-art spatial-temporal graph neural networks without using a pre-defined graph, while DCRNN, STGCN, and MRA-BGCN fully rely on pre-defined graphs. Graph Wavenet proposes a self-adaptive adjacency matrix, but it needs to combine with a pre-defined graph in order to achieve optimal performance. ST-MetaNet employs attention mechanisms to adjust the edge weights of a pre-defined graph. GMAN leverages node2vec algorithm to preserve node structural information while performing attention mechanisms.  When a graph is not defined, these methods cannot model multivariate times series data efficiently.

\begin{table}[tb]
	\centering
	\caption{Baseline comparison under multi-step forecasting for spatial-temporal graph neural networks.} 
	\label{table:multi}
	\resizebox{0.48\textwidth}{!}{
		
		\begin{tabular}{ l  r r r | r r r |r r r}
			\toprule
			\multirow{2}{*}{}   & \multicolumn{3}{c}{Horizon 3} & \multicolumn{3}{c}{Horizon 6}  & \multicolumn{3}{c}{Horizon 12} \\
			\cline{2-4}  \cline{5-7} \cline{8-10}   		    \\[-1em]  
			
			& {\small MAE} & {\small RMSE} & {\small MAPE} & {\small MAE} & {\small RMSE} & {\small MAPE} & {\small MAE} & {\small RMSE} & {\small MAPE}\\
			\midrule
			METR-LA  \\ \cline{1-1}
			\\[-1em]  
			
			DCRNN  & 2.77 & 5.38 & 7.30\% & 3.15 & 6.45 & 8.80\% & 3.60 & 7.60 & 10.50\% \\ 
			STGCN  & 2.88 & 5.74 & 7.62\% & 3.47 & 7.24 & 9.57\% & 4.59 & 9.40 & 12.70\%\\
			Graph WaveNet  & 2.69 & 5.15 & 6.90\% & 3.07 & 6.22 & 8.37\% & 3.53 & 7.37 & 10.01\%\\
			ST-MetaNet & 2.69 & 5.17 & 6.91\% & 3.10 & 6.28 & 8.57\% & 3.59 & 7.52 & 10.63\% \\
			MRA-BGCN  & \textbf{2.67} & \textbf{5.12} & 6.80\% & 3.06 & 6.17 & 8.30\% & 3.49 & 7.30 & 10.00\%\\
			GMAN & 2.77 & 5.48 & 7.25\% & 3.07 & 6.34 & 8.35\% & \textbf{3.40} & \textbf{7.21} & \textbf{9.72\%} \\  		    
			\\[-1em]  
			\hline
			\\[-1em]
			MTGNN & 2.69 &  5.18 & 6.86\% & \textbf{3.05} & \textbf{6.17} & \textbf{8.19\%} & 3.49 & 7.23 & 9.87\%\\
			MTGNN+sampling & 2.76  & 5.34 & \textbf{5.18\%} & 3.11  &  6.32 & 8.47\%  & 3.54 & 7.38 & 10.05\%\\
			\hline \hline 
			PEMS-BAY  \\ \cline{1-1}
			\\[-1em] 
			DCRNN  & 1.38 & 2.95 & 2.90\% & 1.74 & 3.97 & 3.90\% & 2.07 & 4.74 & 4.90\% \\ 
			STGCN  & 1.36 & 2.96 & 2.90\% & 1.81 & 4.27 & 4.17\% & 2.49 & 5.69 & 5.79\%\\
			Graph WaveNet  & 1.30 & 2.74 & \textbf{2.73\%} & 1.63 & 3.70& 3.67\% & 1.95 &  4.52 &  4.63\%\\
			ST-MetaNet & 1.36 & 2.90 & 2.82\% & 1.76 & 4.02 & 4.00\% & 2.20 & 5.06 & 5.45\%\\
			MRA-BGCN  & \textbf{1.29} & \textbf{2.72} & 2.90\% & \textbf{1.61} & \textbf{3.67} & 3.80\% & 1.91 & 4.46 & 4.60\%\\
			GMAN  & 1.34 & 2.82 & 2.81\% & 1.62 & 3.72 & \textbf{3.63\%} & \textbf{1.86} & \textbf{4.32} & \textbf{4.31\%}\\ 		 
			\\[-1em]  
			\hline
			\\[-1em]  
			MTGNN & 1.32 & 2.79& 2.77\% & 1.65& 3.74 & 3.69\% & 1.94 & 4.49 & 4.53\%\\
			MTGNN+sampling & 1.34 & 2.83 & 2.83\% & 1.67 & 3.79& 3.78\% & 1.95 & 4.49 & 4.62\%\\
			\bottomrule
		\end{tabular}
	}
\end{table}

\subsection{Ablation Study}

We conduct an ablation study on the METR-LA data to validate the effectiveness of key components that contribute to the improved outcomes of our proposed model. We name MTGNN without different components as follows:

\begin{itemize}
	\item \textbf{w/o GC}: MTGNN without the graph convolution module. We replace the graph convolution module with a linear layer.
	\item \textbf{w/o Mix-hop}: MTGNN without the information selection step in the mix-hop propagation layer. We pass the outputs of the information propagation step to the next module directly.
	\item \textbf{w/o Inception}: MTGNN without inception in the dilated inception layer. While keeping the same number of output channels, we use a single $1\times 7$ filter only. 
	\item \textbf{w/o CL}: MTGNN without curriculum learning. We train MTGNN without gradually increasing the prediction length.
\end{itemize}

We repeat each experiment 10 times with 50 epochs per repetition and report the average of MAE, RMSE, MAPE with a standard deviation over 10 runs on the validation set in Table \ref{tab:ablation}. The introduction of graph convolution modules significantly improves the results as it enables information flow among isolated but interdependent nodes. The effect of mix-hop is evident as well: it validates that the use of mix-hop is helpful for selecting useful information at each information propagation step in the mix-hop propagation layer. The effect of inception is significant in terms of RMSE, but marginal in terms of MAE. This is because using a single $1\times7$ filter has half more parameters than using a combination of $1\times2,1\times3,1\times5,1\times7$ filters under the condition that the number of output channels for the dilated inception layer remains the same. Lastly, our curriculum learning strategy proves to be effective. It enables our model to converge quickly to an optimum that fits for the easiest task, and  fine-tune parameters step by step as the level of learning difficulty increases.

\begin{table}
	\centering
	\caption{\small Ablation study.}
	\label{tab:ablation}
	\resizebox{0.45\textwidth}{!}{
		\begin{tabular}{l| l|l|l|l|l}
			\toprule
			Methods & MTGNN & w/o GC & w/o Mix-hop & w/o Inception & w/o CL    \\
			\\[-1em]  
			\midrule
			\\[-1em]
			MAE & \textbf{2.7715$\pm$0.0119} &2.8953$\pm$0.0054 & 2.7975$\pm$0.0089 & 2.7772$\pm$0.0100 & 2.7828$\pm$0.0105 \\ 
			\\[-1em]  
			\hline
			\\[-1em]
			RMSE & \textbf{5.8070$\pm$0.0512} &6.1276$\pm$0.0339 & 5.8549$\pm$0.0474 & 5.8251$\pm$0.0429 & 5.8248$\pm$0.0366 \\
			\\[-1em]  
			\hline
			\\[-1em]
			MAPE & \textbf{0.0778$\pm$0.0009} &0.0831$\pm$0.0009 & 0.0779$\pm$0.0009 & 0.0778$\pm$0.0010 & 0.0784$\pm$0.0009 \\ 
			
			\bottomrule 
		\end{tabular}
	}
	
\end{table}

\subsection{Study of the Graph Learning Layer}

To validate the effectiveness of our proposed graph learning layer, we conduct a study which experiments with different ways of constructing a graph adjacency matrix. Table \ref{tab:congraph} shows different forms of $\mathbf{A}$ with experimental results tested on the validation set of the METR-LA data averaged on 10 runs. Predefined-A is constructed by road network distance \cite{li2018diffusion}. Global-A assumes the adjacency matrix is a parameter matrix, which contains $N^2$ parameters. Motivated by \cite{wu2019graph}, Undirected-A and Directed-A are computed by the similarity scores of node embeddings. Motivated by \cite{guo2019attention,shi2019two}, Dynamic-A assumes the spatial dependency at each time step is dependent on its node inputs. 
Uni-directed-A is our proposed method. According to Table \ref{tab:congraph},  our proposed uni-directed-A achieves the lowest mean MAE, RMSE, and MAPE. It improves over predefined-A, undirected-A, and dynamic-A significantly. Our uni-directed-A improves over undirected-A and directed-A marginally in terms of MAE and MAPE but proves to be more robust due to a lower RMSE. 

\begin{figure}
	\begin{subfigure}[b]{0.22\textwidth}   
		\centering 
		\includegraphics[width=\textwidth]{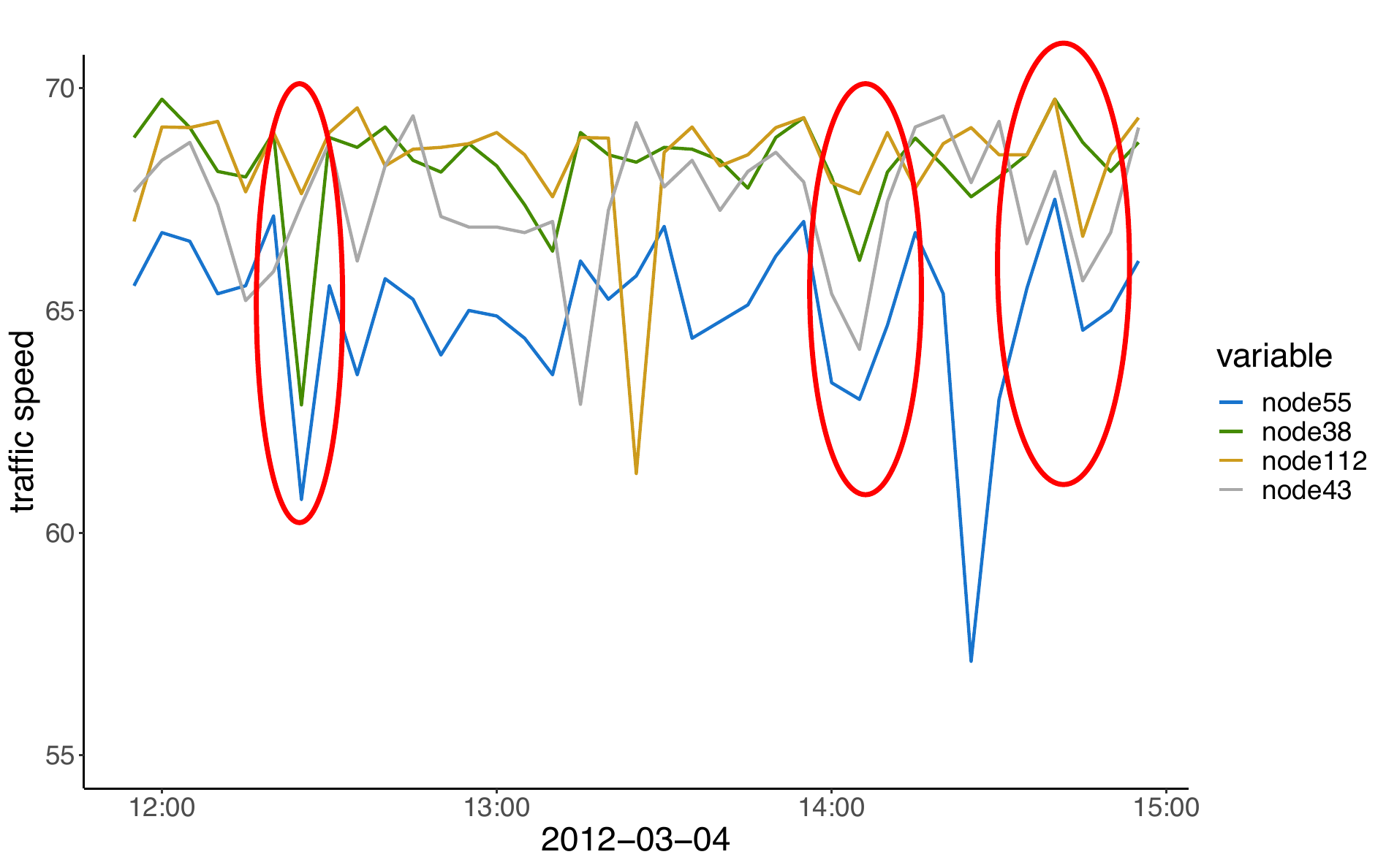}
		\caption[Time series of node 55 and its top-3 neighbors given by the pre-defined $\mathbf{A}$.]
		{{\small Time series of node 55 and its top-3 neighbors given by the pre-defined $\mathbf{A}$.}}    
		\label{fig:prea}
	\end{subfigure}
	\hfill
	\begin{subfigure}[b]{0.22\textwidth}   
		\centering 
		\includegraphics[width=\textwidth]{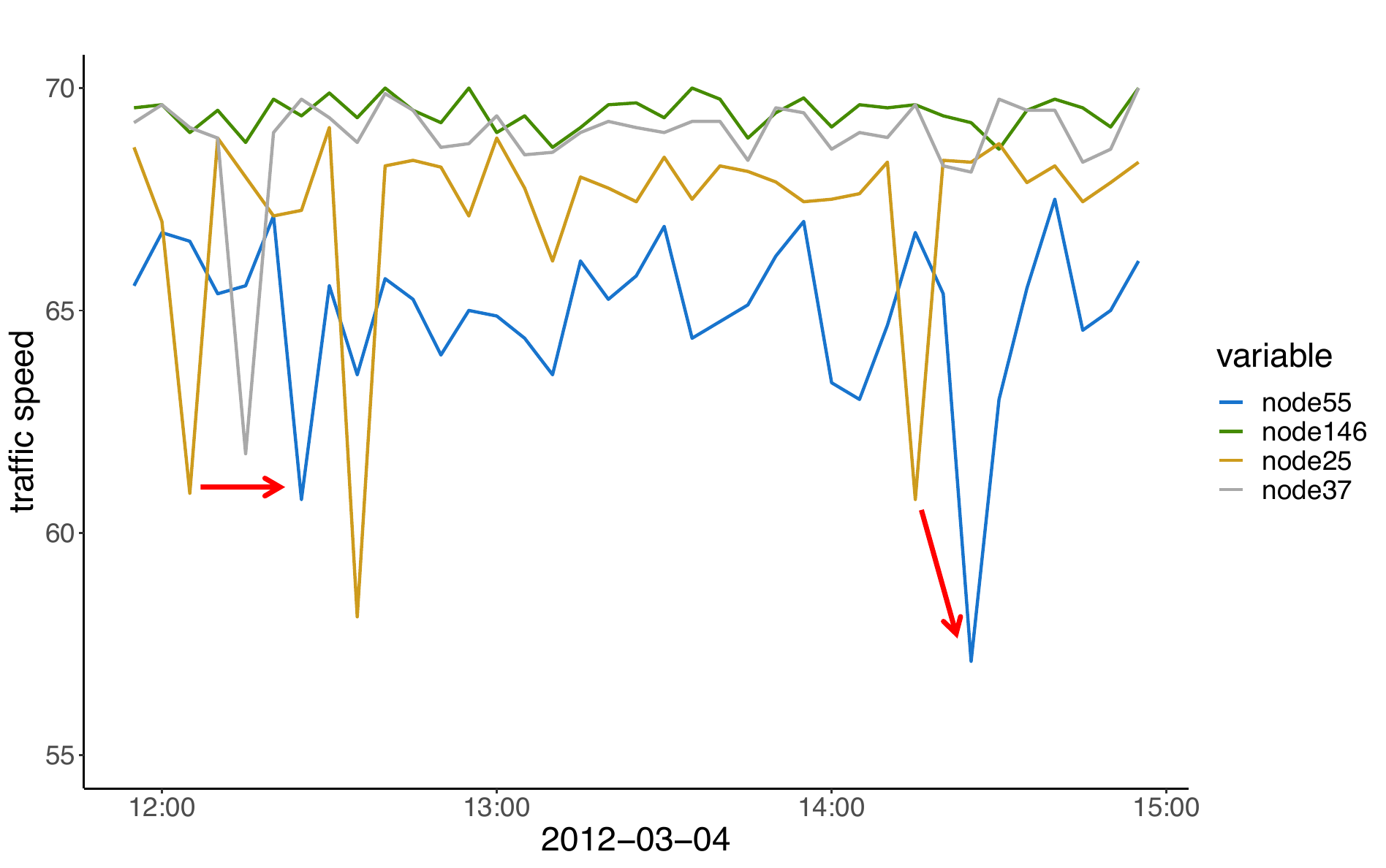}
		\caption[Time series of node 55 and its top-3 neighbors given by the learned $\mathbf{A}$.]
		{{\small Time series of node 55 and its top-3 neighbors given by the learned $\mathbf{A}$.}}    
		\label{fig:leaa}
	\end{subfigure}
	\newline 
	\begin{subfigure}[b]{0.45\textwidth}   
		\centering 
		\includegraphics[width=\textwidth]{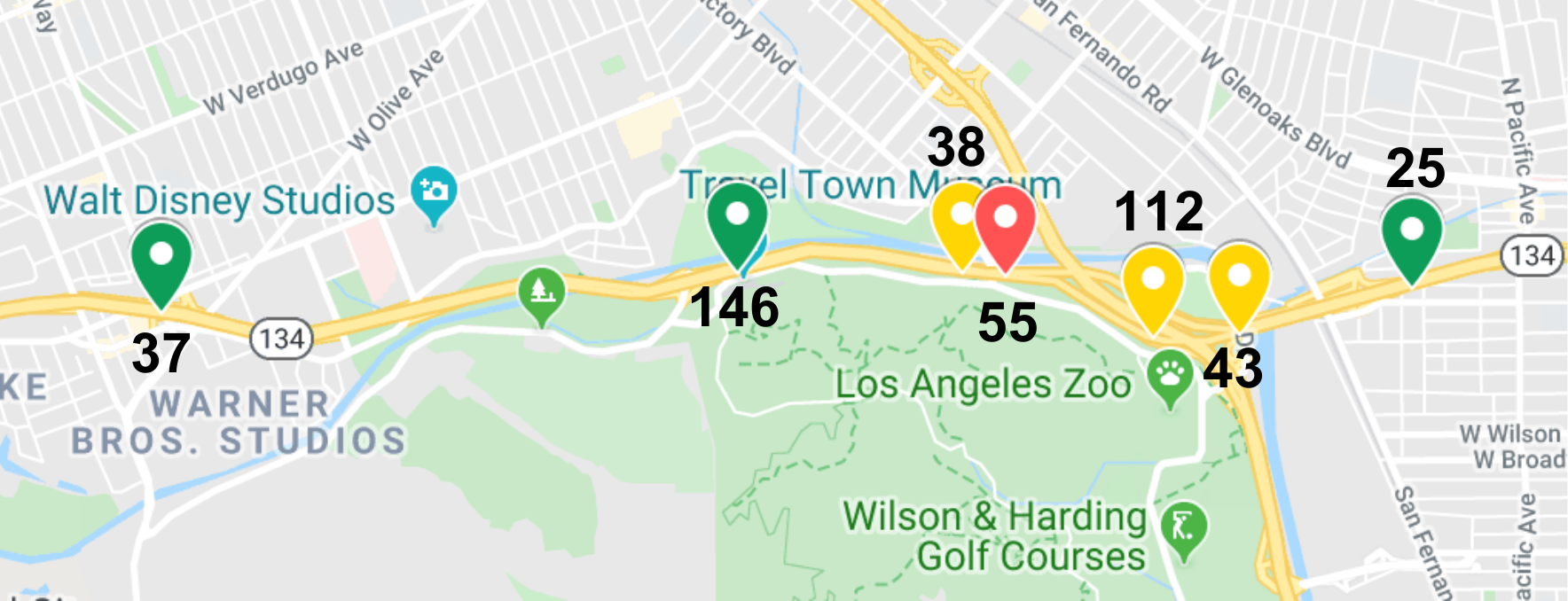}
		\caption[Node locations of node 55 and its neighbors marked on Google Maps. Yellow nodes represent node 55's top-3 neighbors give by the pre-defined $\mathbf{A}$. Green nodes represent node 55's top-3 neighbors give by the learned $\mathbf{A}$.]
		{{\small Node locations of node 55 and its neighbors marked on Google Maps. Yellow nodes represent node 55's top-3 neighbors given by the pre-defined $\mathbf{A}$. Green nodes represent node 55's top-3 neighbors given by the learned $\mathbf{A}$.}}    
		\label{fig:nodeloc}
	\end{subfigure}
	\caption[Case study]
	{\small Case study} 
	\label{fig:case}
\end{figure}

We further investigate the learned graph adjacency matrix via a case study. In Figure \ref{fig:prea}, we plot the raw time series of node 55 and its pre-defined top-3 neighbors. In Figure \ref{fig:leaa}, we chart the raw time series of node 55 and its learned top-3 neighbors. Figure \ref{fig:nodeloc} shows the geo-location of these nodes, with green nodes representing the central node's learned top-3 neighbors and yellow nodes representing the central node's pre-defined top-3 neighbors. We observe that the central node's pre-defined top-3 neighbors are much closer to the node itself on the map. As a result, their time series are more correlated simultaneously, as shown by the red circles in Figure \ref{fig:prea}.  On the contrary,  the central node's learned top-3 neighbors distribute further away from it but still lie on the same road it follows. According to Figure \ref{fig:leaa}, time series of the learned top-3 neighbors are more capable of indicating extreme traffic conditions of the central node in advance.  


\begin{table}
	\centering
	\caption{Comparison of different graph learning methods.}
	\label{tab:congraph}
	\resizebox{0.45\textwidth}{!}{
		\begin{tabular}{ll|l|l|l}
			\toprule
			\\[-1em]
			Methods & Equation & MAE & RMSE & MAPE\\
			\\[-1em]  
			\midrule
			\\[-1em]
			Pre-defined-A & - & 2.9017$\pm$0.0078 & 6.1288$\pm$0.0345 & 0.0836$\pm$0.0009\\
			\\[-1em]  
			\hline
			\\[-1em]
			Global-A & $\mathbf{A}=ReLU(\mathbf{W})$ & 2.8457$\pm$0.0107 & 5.9900$\pm$0.0390 & 0.0805$\pm$0.0009 \\ 
			\\[-1em]  
			\hline
			\\[-1em]
			Undirected-A  & $\mathbf{A}=ReLU(tanh(\alpha (\mathbf{M}_1\mathbf{M}_1^T)))$  & 2.7736$\pm$0.0185 & 5.8411$\pm$0.0523 & 0.0783$\pm$0.0012\\ 
			\\[-1em]  
			\hline
			\\[-1em]
			Directed-A  & $\mathbf{A}=ReLU(tanh(\alpha (\mathbf{M}_1\mathbf{M}_2^T)))$  & 2.7758$\pm$0.0088 & 5.8217$\pm$0.0451 & 0.0783$\pm$0.0006\\ \\[-1em]  
			\hline
			\\[-1em]
			Dynamic-A & $\mathbf{A}_t=SoftMax(tanh(\mathbf{X}_t\mathbf{W}_1)tanh(\mathbf{W}_2^T\mathbf{X}_t^T))$ & 2.8124$\pm$0.0102 & 5.9189$\pm$0.0281
			& 0.0794$\pm$0.0008 \\ \\[-1em]  
			\hline
			\\[-1em]
			Uni-directed-A (ours) & $\mathbf{A}=ReLU(tanh(\alpha (\mathbf{M}_1\mathbf{M}_2^T-\mathbf{M}_2\mathbf{M}_1^T)))$ & \textbf{2.7715$\pm$0.0119} & \textbf{5.8070$\pm$0.0512} & \textbf{0.0778$\pm$0.0009} \\
			\\[-1em]  
			\bottomrule
		\end{tabular}
	}
\end{table}

\section{Conclusions}
In this paper, we introduce a novel framework for multivariate time series forecasting.  To the best of our knowledge, we are the first to address the multivariate time series forecasting problem via a graph-based deep learning approach. We propose an effective method to exploit the inherent dependency relationships among multiple time series. Our method demonstrates superb performance in a variety of multivariate time series forecasting tasks and  opens a new door to use GNNs to handle diverse non-structural data. 


\begin{acks}
	This work was supported in part by the Australian Research Council (ARC) under Grant LP160100630, LP180100654 and DE190100626. 
\end{acks}

%
\bibliographystyle{ACM-Reference-Format}
\bibliography{main}

\appendix
\section{APPENDIX: REPRODUCIBILITY}
In this section, we provide the details of our implementation for reproducibility. Our source codes \footnote{https://github.com/nnzhan/MTGNN} are publicly available.



\subsection{Complexity Analysis}
We analyze the time complexity of the main components of the proposed model MTGNN, which is summarized in Table \ref{tab:time}. The time complexity of the graph learning layer is $(O(Ns_1s_2+N^2s_2))$ where $N$ denotes the number of nodes,  $s_1$ represents the dimension of node input feature vectors, and $s_2$ represents the dimension of node hidden feature vectors. Treating $s_1$ and $s_2$ as constants, the time complexity of the graph learning layer becomes $O(N^2)$.  It is attributed to the pairwise computation of node hidden feature vectors.  The graph convolution module incurs $O(K(Md_1+Nd_1d_2)$ time complexity,  where $K$ is the propagation depth, $N$ is the number of nodes, $d_1$ denotes the input dimension of node states, $d_2$ denotes the output dimension of node states.  Regarding $K$, $d_1$ and $d_2$ as constants, the time complexity of the graph convolution module turns to $O(M)$. This result comes from the fact that in the information propagation step, each node receives information from its neighbors and the sum of the number of neighbors of each node exactly equals the number of edges.  The time complexity of the temporal convolution module equals to $O(Nlc_ic_o/d)$, where $l$ is the input sequence length, $c_i$ is the number of input channels, $c_o$ is the number of output channels, and $d$ is the dilation factor. The time complexity of the temporal convolution module mainly depends on $N\times l$, which is the size of the input feature map.

\begin{table}[H]
	\centering
	\resizebox{0.45\textwidth}{!}{
		\begin{tabular}{l | l}
			\toprule
			Components & Time Complexity    \\
			\\[-1em]  
			\midrule
			\\[-1em]
			Graph Learning Layer & $O(Ns_1s_2+N^2s_2)$ \\ 
			\\[-1em]  
			\hline
			\\[-1em]
			Graph Convolution Module & $O(K(Md_1+Nd_1d_2)$\\
			\\[-1em]  
			\hline
			\\[-1em]
			Temporal Convolution Module & $O(Nlc_ic_o/d)$\\ 
			
			\bottomrule 
		\end{tabular}
	}
	\caption{Time Complexity Analysis}
	\label{tab:time}
\end{table}

\subsection{Data}\label{sec:dataset}
In Table \ref{tb:data-stats}, we summarize statistics of benchmark datasets. Details of these datasets are introduced below.

\subsubsection{Single-step forecasting}
\begin{itemize}
    \item Traffic: the traffic dataset from the California Department of Transportation contains road occupancy rates measured by 862 sensors in San Francisco Bay area freeways during 2015 and 2016.
    \item Solar-Energy: the solar-energy dataset from the National Renewable Energy Laboratory contains
    the solar power output collected from 137 PV plants in Alabama State in 2007.
    \item Electricity:  the electricity dataset from the UCI Machine Learning Repository contains electricity consumption for 321 clients from 2012 to 2014.
    \item Exchange-Rate: the exchange-rate dataset contains the daily exchange rates of eight foreign countries including Australia, British, Canada, Switzerland, China, Japan, New Zealand, and Singapore ranging from 1990 to 2016.
\end{itemize}
Following \cite{lai2018modeling}, we split these four datasets into a training set (60\%), validation set (20\%), and test set (20\%) in chronological order. The input sequence length is 168 and the output sequence length is 1.
Models are trained independently to predict the target future step (horizon) 3, 6, 12, and 24.

\subsubsection{Multi-step forecasting}
\begin{itemize}
    \item METR-LA: the METR-LA dataset from the Los Angeles Metropolitan Transportation Authority contains average traffic speed measured by 207 loop detectors 
    on the highways of Los Angeles County ranging from Mar 2012 to Jun 2012.
    \item PEMS-BAY: the PEMS-BAY dataset from California Transportation Agencies (CalTrans) contains average traffic speed measured by 325 sensors 
    in the Bay Area ranging from Jan 2017 to May 2017.
\end{itemize}
Following \cite{li2018diffusion}, we split these two datasets into a training set (70\%), validation set (20\%), and test set (10\%) in chronological order. The input sequence length is 12, and the target sequence contains the next 12 future steps.
The time of the day is used as an auxiliary feature for the inputs. For the selected baseline methods, the pairwise road network distances are used as the pre-defined graph structure.

\subsection{Experimental Setup}\label{sec:expsetup}
We repeat the experiment 10 times and report the average value of evaluation metrics. The model is trained by the Adam optimizer with gradient clip 5. The learning rate is 0.001. The l2 regularization penalty is 0.0001. Dropout with 0.3 is applied after each temporal convolution module. Layernorm is applied after each graph convolution module. The depth of the mix-hop propagation layer is set to 2. The retain ratio from the mix-hop propagation layer is set to $0.05$. 
The saturation rate of the activation function from the graph learning layer is set to $3$.
The dimension of node embeddings is 40. Other hyper-parameters are reported according to different tasks.  

\subsubsection{Single-step forecasting}
We use 5 graph convolution modules and 5 temporal convolution modules with the dilation exponential factor 2. The starting $1\times1$ convolution has 1 input channel and 16 output channels. The graph convolution module and the temporal convolution modules both have 16 output channels. The skip connection layers all have 32 output channels. The first layer of the output module has 64 output channels and the second layer of the output module has 1 output channel. The number of training epochs is 30. For Traffic, Solar-Energy, and Electricity, the number of neighbors for each node is 20. For Exchange-Rate, the number of neighbors for each node is 8. The batch size is set to 4.  For the MTGNN+sampling model, we split the nodes of a graph into three partitions randomly with a batch size of 16. Following \cite{lai2018modeling}, we use RSE and CORR as evaluation metrics.

\subsubsection{Multi-step forecasting}
We use 3 graph convolution modules and 3 temporal convolution modules with the dilation exponential factor 1. The starting $1\times1$ convolution has 2 input channels and 32 output channels. The graph convolution module and the temporal convolution module both have 32 output channels. The skip connection layers all have 64 output channels. The first layer of the output module has 128 output channels and its second layer has 12 output channels. The number of neighbors for each node is 20. The number of training epochs is 100. The batch size is set to 64. Following \cite{li2018diffusion}, we use MAE, RMSE , and MAPE as evaluation metrics.

\subsection{Parameter Study}
We conduct a parameter study on eight core hyper-parameters which influence the model complexity of MTGNN. We list these hyper-parameters as follows: Number of layers, the number of temporal convolution modules, ranges from 1 to 6. Number of filters, the number of output channels for temporal convolution modules and graph convolution modules, ranges from 4 to 128. Number of neighbors, the parameter $k$ in Equation \ref{eq:argtop}, ranges from 10 to 60. 
Saturation rate, the parameter $\alpha$ in Equation \ref{eq:m1}, \ref{eq:m2}, and \ref{eq:adp}, ranges from 0.5 to 5. Retain ratio of mix-hop propagation layer, the parameter $\beta$ in Equation \ref{eq:ppnp}, ranges from 0 to 0.8. Depth of mix-hop propagation layer, the parameter $K$ in Equation \ref{eq:mix}, ranges from 1 to 6. 

\begin{figure}
        \begin{subfigure}[b]{0.23\textwidth}   
            \centering 
            \includegraphics[width=\textwidth]{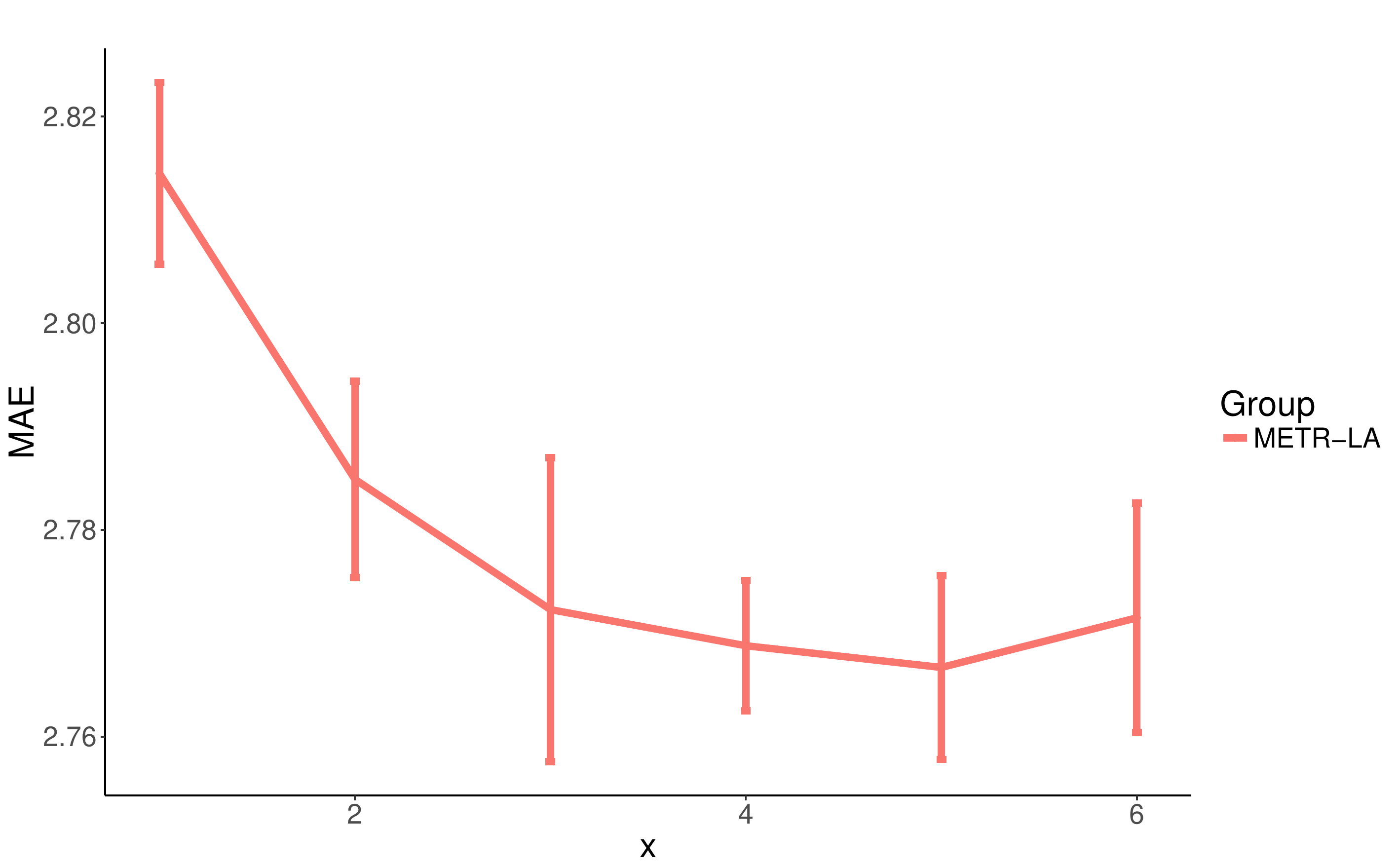}
            \caption[Number of layers]%
            {{\small Number of layers}}    
            \label{fig:6a}
        \end{subfigure}
        \hfill
        \begin{subfigure}[b]{0.23\textwidth}   
            \centering 
            \includegraphics[width=\textwidth]{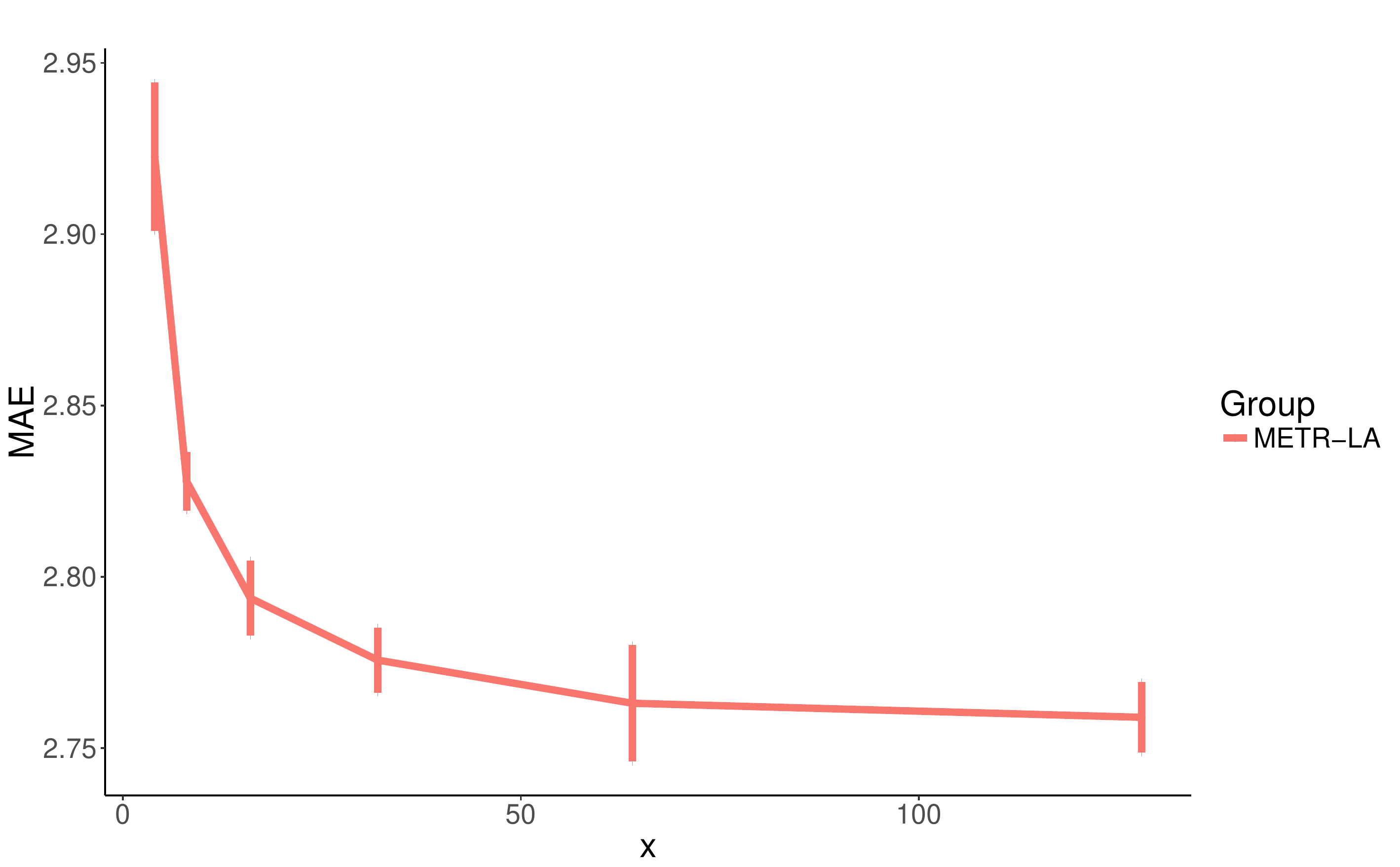}
            \caption[Number of filters]%
            {{\small Number of filters}}    
            \label{fig:6b}
        \end{subfigure}
        \newline
        \begin{subfigure}[b]{0.23\textwidth}   
            \centering 
            \includegraphics[width=\textwidth]{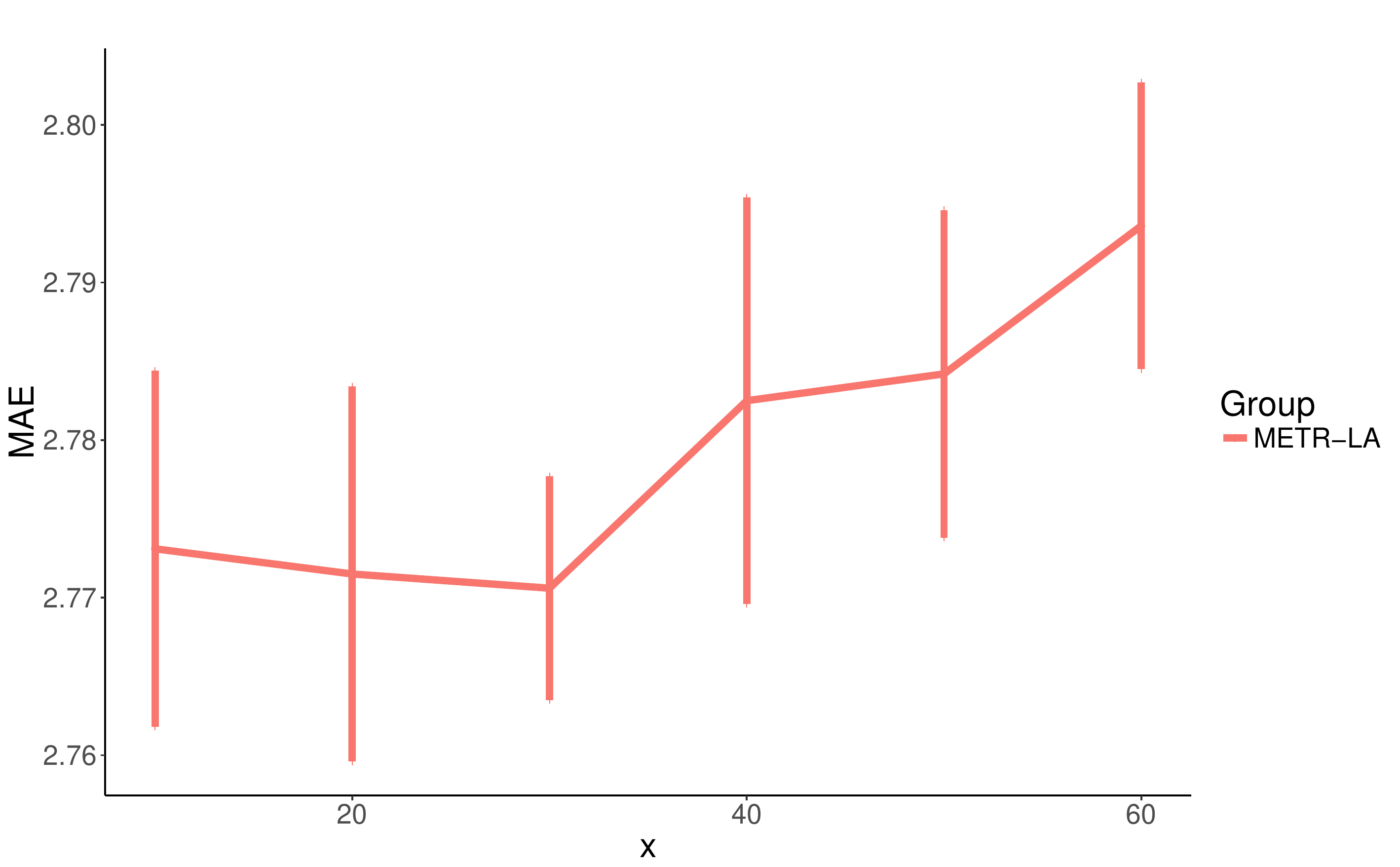}
            \caption[Number of neighbors]%
            {{\small Number of neighbors}}    
            \label{fig:6c}
        \end{subfigure}
        \hfill
        \begin{subfigure}[b]{0.23\textwidth}   
            \centering 
            \includegraphics[width=\textwidth]{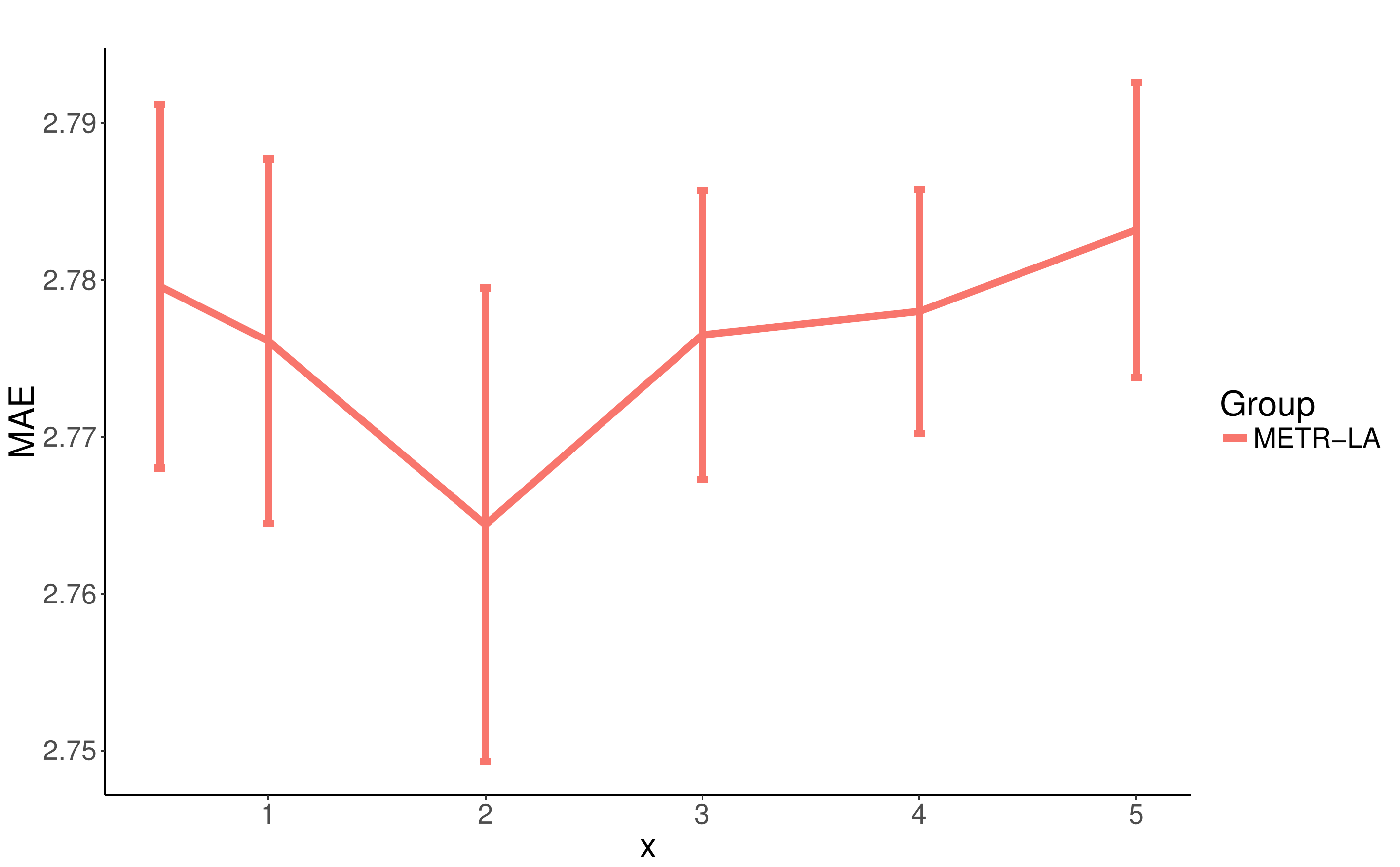}
            \caption[Saturation rate]%
            {{\small Saturation rate}}    
            \label{fig:6d}
        \end{subfigure}
        \newline
        \begin{subfigure}[b]{0.23\textwidth}   
            \centering 
            \includegraphics[width=\textwidth]{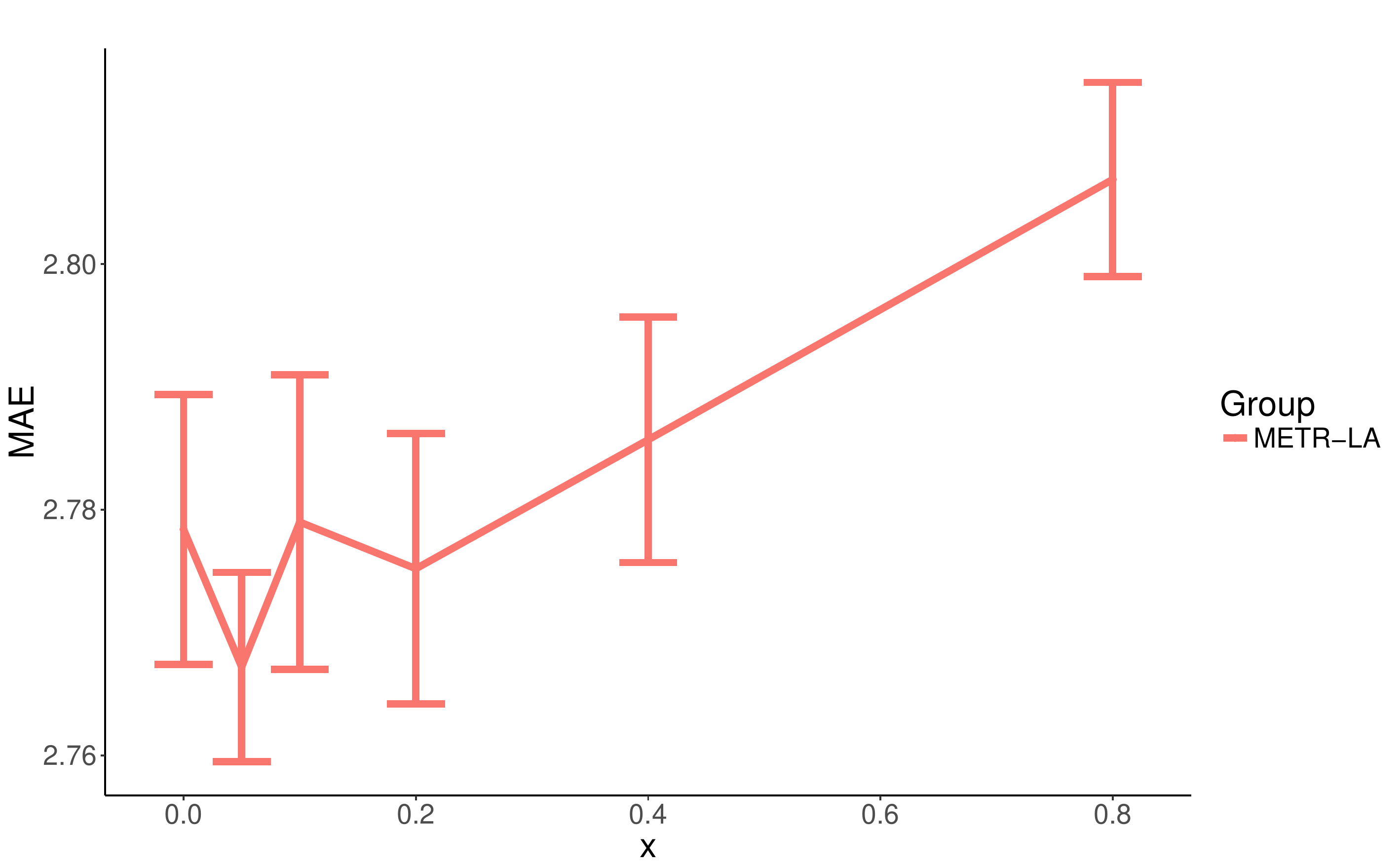}
            \caption[Retain ratio]%
            {{\small Retain ratio}}    
            \label{fig:6e}
        \end{subfigure}
        \hfill
        \begin{subfigure}[b]{0.23\textwidth}   
            \centering 
            \includegraphics[width=\textwidth]{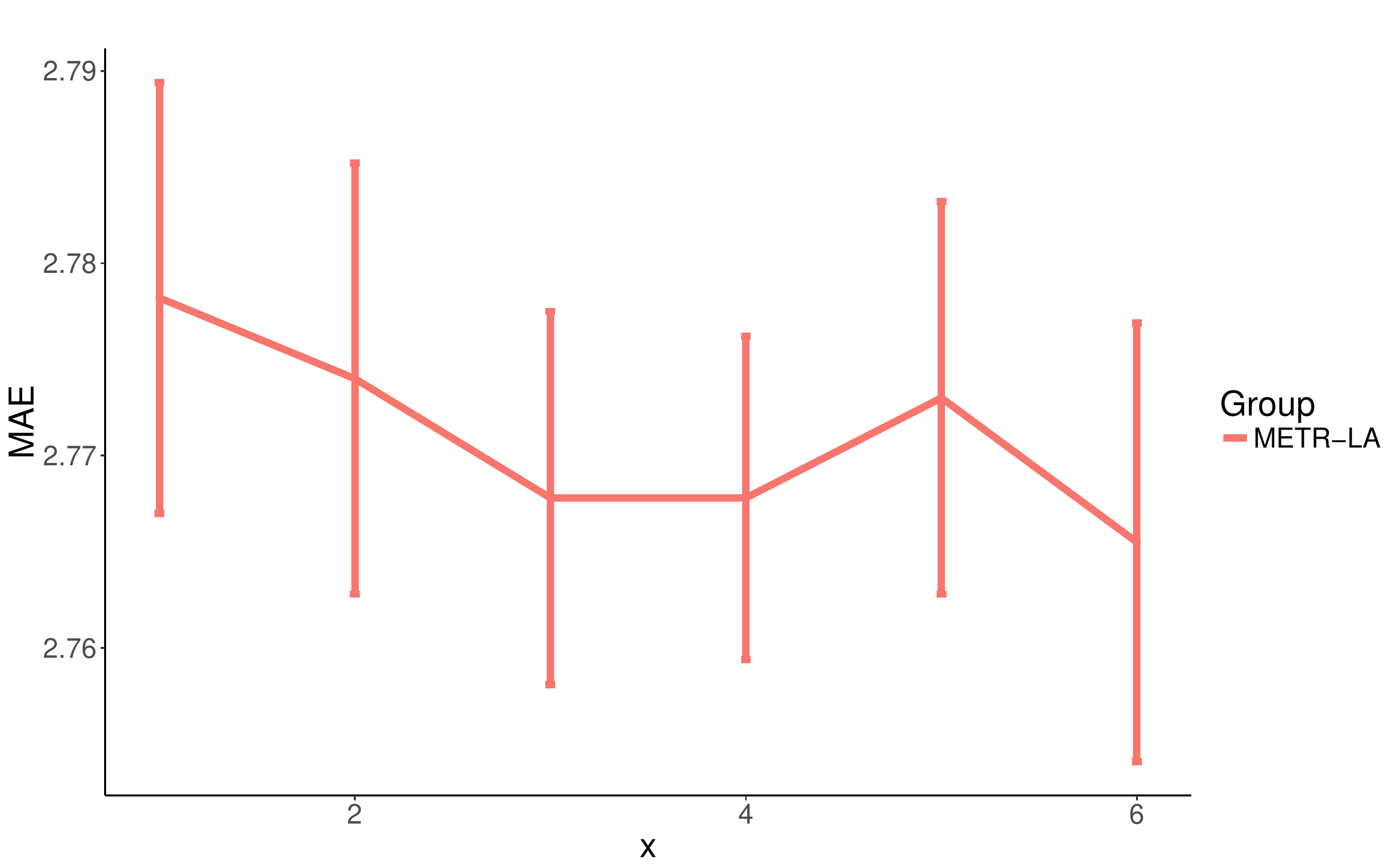}
            \caption[Depth of propagation]%
            {{\small Depth of propagation}}    
            \label{fig:6f}
        \end{subfigure}
        \caption[Parameter Study]
        {\small Parameter Study} 
        \label{fig:para}
\end{figure}

We repeat each experiment 10 times with 50 epochs each time and report the average of MAE with a standard deviation over 10 runs on the validation set. We change the parameter under investigation and fix other parameters in each experiment.  Figure \ref{fig:para} shows the experimental results of our parameter study. As shown in Figure \ref{fig:6a} and Figure \ref{fig:6b}, increasing the number of layers and filters enhances our model's expressive capacity, while reducing the MAE loss. Figure \ref{fig:6c} shows that a small number of neighbors gives better results. It is possibly because a node may only depend on a limited number of other nodes, and increasing its neighborhood merely introduces noises to the model. The model performance is not sensitive to the saturation rate, as shown in Figure \ref{fig:6d}. However, a large saturation rate can impose values of the adjacency matrix produced by the graph learning layer approach to 0 or 1. As shown in Figure \ref{fig:6e}, a high retain ratio degrades the model performance significantly. We think it is because by default the propagation depth of the mix-hop propagation layer is set to $2$, and as a result, keeping a high proportion of root information constrains a node from exploring its neighborhood. Figure \ref{fig:6f} shows that it is enough to propagate node information with 2 or 3 steps. With the increase of the depth of propagation, the proposed mix-hop propagation layer does not suffer from the over-smoothing problem incurred by information aggregation. With the depth of propagation equal to 6, it has the lowest mean MAE with a larger variation.

\end{document}